\documentclass{article}
\usepackage{textcomp}

\usepackage[main, final]{neurips_2026}

\usepackage[utf8]{inputenc}
\usepackage[T1]{fontenc}
\usepackage{graphicx}
\usepackage{gensymb}

\usepackage{algorithm}
\usepackage{algpseudocode}
\usepackage{hyperref}
\usepackage{url}
\usepackage{microtype}
\usepackage{booktabs}
\usepackage{multirow}

\usepackage{amsmath,amssymb,amsthm,amsfonts}
\usepackage{nicefrac}

\usepackage{xcolor}
\usepackage{float}
\usepackage{placeins}

\usepackage{caption}
\usepackage{subcaption}
\usepackage{makecell}

\usepackage{enumitem}
\usepackage{pifont}
\usepackage{needspace}

\newtheorem{proposition}{Proposition}

\newtheorem{definition}{Definition}
\newtheorem{remark}{Remark}

\title{When Does Subspace Direction Matter for LoRA?\\
Regime Analysis of the Magnitude Principle in Few-Shot Adaptation}

\author{%
  \textbf{Nischal Subedi} \\
  Univ. of Delaware \\
  \texttt{nsubedi@udel.edu} \\
  \And
  \textbf{Cencheng Shen} \\
  Microsoft Research \\
  \texttt{c.shen@microsoft.com} \\
  \And
  \textbf{Peng Zhao} \\
  Univ. of Delaware \\
  \texttt{pzhao@udel.edu} \\
}

\begin{document}

\maketitle

\begin{abstract}
We show that the importance of a LoRA adapter's initial input-subspace
direction is regime-dependent, and we characterize this dependence in
theory and experiment. This refines the recent view that spectral
initializers help through update magnitude alone. Direction matters most
when labeled data are scarce relative to the representation dimension
and the activation subspace used for initialization captures supervised
task structure. In this regime, the adapter's row span remains sensitive
to its starting point, making task-aligned initialization especially
useful. A spiked-covariance perturbation analysis characterizes the
finite-sample stability of the estimated subspace. A first-step analysis
of LoRA training explains why the subspace's alignment with the task
matters and motivates an alignment coefficient $\rho$ that closely tracks
gains across classification tasks. We test this account with PivotLoRA, which initializes adapters
from activation PCA at prediction-relevant positions and applies one
output-preserving orthogonal pivot early in training. In few-shot
BERT-base classification, PivotLoRA achieves the best average rank among
13 methods and improves over vanilla LoRA in 29 of 30 settings, even
though it starts from a smaller input-factor initialization norm.
Additional classification experiments extend the evaluation to 7B and
8B Llama decoders. Classification gains diminish with more labeled data,
consistent with the proposed regime dependence. A targeted extension to
completion-only reasoning also shows improvements on GSM8K and
StrategyQA. These findings identify conditions under which initialization
direction matters, with subspace--task alignment offering a low-cost
diagnostic before training for classification.
\end{abstract}

\section{Introduction}

Low-Rank Adaptation~\citep{hu2022lora} efficiently adapts pretrained models by
parameterizing weight updates with low-rank factors. The input factor
determines the feature subspace through which adaptation initially proceeds.
Existing initialization methods derive directions for this subspace from
singular components of pretrained
weights~\citep{meng2024pissa,wang2025milora}, task
gradients~\citep{wang2024loraga,zhang2025loraone}, or activation
statistics~\citep{paischer2025eva,yang2024corda}. Recent work attributes gains
from pretrained-weight spectral initialization primarily to amplified
update magnitudes instead of the selected singular directions
themselves~\citep{zhang2025loram}. This leaves open when initialization
direction materially affects adaptation. We study this question primarily
in classification, with a targeted extension to generation, by applying
PCA to frozen-model activations from the adaptation set at positions
associated with the supervised prediction. These include the
\texttt{[CLS]} position in encoder classification, the mean over input
positions in decoder classification, where the label describes the whole
sequence, and hidden states that predict supervised response tokens in
completion-only generation.

Our central claim is that the value of initialization direction is
regime-dependent. Direction matters most when labeled data are scarce
relative to the representation dimension and the estimated subspace
captures supervised task structure. A spiked-covariance Davis--Kahan
analysis shows what limits such an estimate, since its error depends on
the effective dimension-to-sample ratio and the spectral gap. The same
qualitative pattern appears in synthetic experiments and transformer
activations, where the ambient ratio $d/N$ provides a simple proxy for
the dimensional regime. For BERT-base, $d=768$ and $N=16$ gives
$d/N=48$; for 7B and 8B decoders with $d=4096$, the high-dimensional
regime persists at $N=64$. A large $d/N$ is not sufficient, however,
because the selected subspace must also contain structure relevant to
the supervised objective.

The role of alignment also appears in the first useful LoRA update.
Standard LoRA initializes the output factor with $B=0$, so initially
$\Delta W=0$ and the gradient with respect to the input factor $A$ is
zero. The first nonzero adapter signal enters through the projected
features $Ah(x)$ in the update to $B$. Under a class-conditional model,
the squared norm of the initial population gradient depends on the
supervised structure captured by the row span of $A$. We quantify this
structure with an alignment coefficient $\rho$, defined as the fraction
of between-class scatter captured by a candidate subspace. For BERT-base
with rank $r=8$, a random subspace captures approximately
$r/d\approx1\%$ of this signal in expectation, whereas prediction-position
PCA captures up to $84\%$ on high-alignment tasks. Across seven
classification tasks, $\rho$ tracks PivotLoRA's shared-protocol gain
with Pearson $r=0.93$ and exact permutation $p=0.002$. The coefficient
$\rho$ thus offers a low-cost measure of subspace--task alignment for
classification.

Motivated by this analysis, we propose \textbf{PivotLoRA}. The method
initializes the LoRA input factor from prediction-relevant activation PCA
and allows early task gradients to refine its span. As the factors evolve,
however, the input basis can acquire unequal scales and correlations that
affect subsequent optimization. We therefore separate learning the span
from normalizing its coordinates. After $K$ epochs, PivotLoRA performs one
output-preserving orthogonal pivot. Under a full-row-rank condition, the
pivot replaces the learned input basis with an orthonormal basis for the
same row span and re-encodes the output factor to preserve the adapter
update. This retains the task-adapted directions while resetting
$AA^\top$ to the identity, allowing training to continue in normalized
coordinates without changing the model function at the pivot.

Our primary evaluation focuses on few-shot classification, covering
30 BERT-base settings and 13 methods, with additional classification
experiments on Llama-2-7B and Llama-3.1-8B. PivotLoRA ranks first overall
and improves over vanilla LoRA in 29 of 30 primary BERT settings, with
the largest shared-protocol gains concentrated in low-sample,
high-alignment tasks. PivotLoRA achieves these gains despite starting
from a smaller initialization norm than vanilla LoRA. Separately,
restoring LoRAM's intended initialization magnitude does not eliminate
its performance deficit relative to vanilla LoRA. In full-data
classification, the gains from initialization direction diminish,
consistent with the proposed dependence on labeled sample size.
Beyond classification, we conduct a targeted evaluation of the activation
construction for generation. GPT-2 summarization shows that a large $d/N$ alone is not enough, as PivotLoRA is competitive with LoRA there but not decisively better. We also evaluate
response-predictor activation initialization in completion-only adaptation
with Llama-3.1-8B. Under a matched protocol, this extension improves
generated-answer accuracy on GSM8K and StrategyQA by 4.5 and 3.6
percentage points, respectively.

\textbf{Contributions.}
First, we identify a regime in which LoRA initialization direction matters,
combining a high dimension-to-sample ratio with strong subspace--task
alignment. Second, we support this view through a spiked-covariance
Davis--Kahan analysis, a local first-step LoRA analysis, and validation
on synthetic and transformer activations. Third, we propose
\textbf{PivotLoRA}, which combines prediction-relevant activation-subspace
initialization with one output-preserving orthogonal reparameterization.
Fourth, we introduce $\rho$ as a frozen-forward-pass diagnostic for
classification and evaluate PivotLoRA primarily through few-shot
classification across encoder and decoder models, complemented by
full-data and summarization controls and a targeted extension to
generated-answer reasoning.
\vspace{-1em}

\section{Related Work}
\label{sec:related}

\paragraph{LoRA initialization.}
Initializers for LoRA~\citep{hu2022lora} draw the adapter subspace from three
sources. Weight-based methods use singular components of the pretrained
matrix, either the principal ones~\citep{meng2024pissa} or the minor
ones~\citep{wang2025milora}. Gradient-based methods take the SVD of a task
gradient~\citep{wang2024loraga,zhang2025loraone} or of a Fisher-weighted
variant~\citep{zhang2026lorada}. Activation-based methods estimate directions
from downstream inputs, either from all-token covariances~\citep{paischer2025eva}
or by modulating the weight decomposition with activation
statistics~\citep{yang2024corda}. PivotLoRA belongs to the third family but
estimates its subspace at prediction-relevant positions, and we test when direction matters beyond scale.

\paragraph{Magnitude and optimization accounts.}
\citet{zhang2025loram} argue that the gains of weight-spectral initializers
largely reflect amplified update magnitude, and \citet{lee2026learningrate}
find that vanilla LoRA matches many of its variants once each method's learning rate is tuned. Differential factor learning rates~\citep{hayou2024loraplus} change the optimization without changing the adapter subspace, whereas periodic re-initialization or resampling~\citep{yang2025srlora,hao2024flora} replaces directions during training. The PivotLoRA pivot instead adds no
new direction and preserves the adapter update exactly, changing only the
coordinates in which optimization continues. Our analysis complements these
accounts by characterizing a regime in which direction matters, with controls
that separate direction from learning-rate allocation, together with
initialization-norm comparisons (extended review in Appendix~\ref{app:related}).

\section{When Does Initialization Direction Matter?}
\label{sec:theory}

We argue that LoRA initialization direction matters most in a regime that
combines a high dimension-to-sample ratio with strong alignment
between the estimated activation subspace and supervised task structure. A
spiked-covariance perturbation analysis characterizes the finite-sample
stability of activation PCA, while an alignment coefficient $\rho$ quantifies
the supervised structure captured by the resulting subspace. The analysis is
most direct for classification settings in which predictions are formed from
a designated representation, such as the \texttt{[CLS]} representation in
BERT.

\paragraph{Notation.}
Let $N$ denote the total number of labeled examples in the few-shot training
set, sampled approximately class-balanced across $C$ classes, so each class has
roughly $n_c \approx N/C$ examples. PivotLoRA estimates its PCA subspace from
the global target-token covariance over all $N$ examples. The between-class
signal used by the alignment diagnostic depends on class means estimated from
approximately $n_c$ examples each. Since $C$ is fixed within a task, replacing
$N$ by $n_c$ changes the dimension-to-sample ratio only by a task-dependent
constant. We therefore express the theoretical regime in terms of $d_{\mathrm{eff}}/N$
and use $d/N$ empirically as a simple architecture-level proxy, not as the
literal quantity in the perturbation bound.

\paragraph{Subspace estimation.}
We model target-token activations at an adapted layer as draws from a spiked
covariance distribution
\begin{equation}
\label{eq:spiked}
  \Sigma
  =
  \sum_{i=1}^{r^\ast}
  (\lambda_i-\sigma^2)v_i v_i^\top
  +
  \sigma^2 I,
\end{equation}
where $\lambda_i>\sigma^2$, the vectors $v_i$ are orthonormal signal
directions, and $\sigma^2 I$ is an isotropic noise floor. This model is
idealized, but it approximates the variance concentration observed in transformer activations. Appendix~\ref{app:spiked-validation} shows that the top eight
components of BERT-base \texttt{[CLS]} activations capture 38--46\% of the
total variance.

Under a sub-Gaussian sampling assumption, combining Davis--Kahan perturbation
theory~\citep{davis1970rotation,yu2015useful} with effective-rank covariance
concentration gives the regime-level scaling
\begin{equation}
\label{eq:dk}
  \mathbb{E}
  \left[
    \|\sin\Theta(\hat V,V^\ast)\|_F
  \right]
  \lesssim
  \min
  \left\{
    \sqrt{r^\ast},
    \sqrt{r^\ast}\,
    \frac{\|\Sigma\|_{\mathrm{op}}}{\delta}
    \left(
      \sqrt{\frac{d_{\mathrm{eff}}}{N}}
      +
      \frac{d_{\mathrm{eff}}}{N}
    \right)
  \right\},
\end{equation}
where $\delta=\lambda_{r^\ast}-\sigma^2$ is the eigengap and

\[
  d_{\mathrm{eff}}
  =
  \frac{\operatorname{Tr}(\Sigma)}
       {\|\Sigma\|_{\mathrm{op}}}
\]
is the effective rank. The quantity $d_{\mathrm{eff}}$ reflects activation anisotropy more faithfully than the ambient dimension $d$.
Appendix~\ref{app:proof-dk} gives the derivation. The bound provides stronger
control when $N$ is large relative to $d_{\mathrm{eff}}$ and the eigengap is
large. When either weakens, the estimated subspace becomes more sensitive to finite-sample variation.

\begin{figure}[t]
  \centering
  \includegraphics[width=0.92\linewidth]{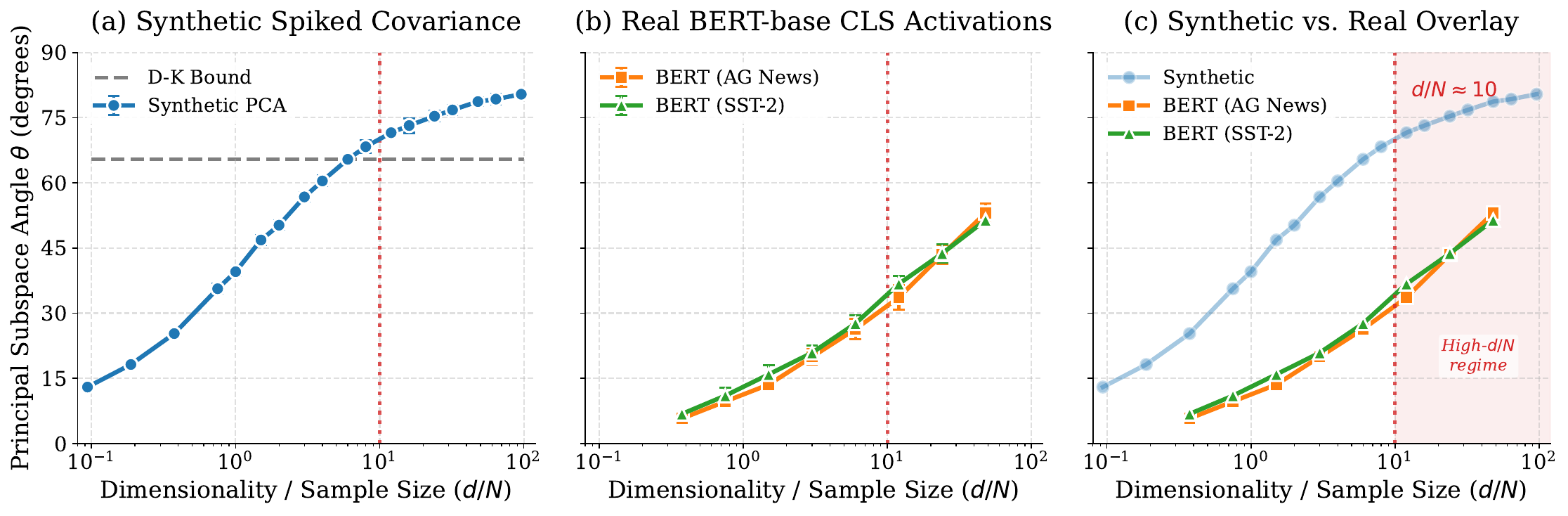}
  \caption{PCA estimation error vs.\ $d/N$. \textbf{(a)} Synthetic spiked
covariance. \textbf{(b)} BERT-base \texttt{[CLS]} activations on AG~News and
SST-2. \textbf{(c)} Overlay. The shaded region marks the empirical high-$d/N$
regime associated with the largest direction-aware gains in our experiments.
Error bars show $\pm 1$ std.}
  \label{fig:spiked}
\vspace{-1.0em}
\end{figure}

\paragraph{Empirical validation.}
Figure~\ref{fig:spiked} evaluates this prediction using PCA estimation error
from both a synthetic spiked model and real BERT-base activations. For the
real-activation experiment, we compute PCA on few-shot \texttt{[CLS]}
embeddings from AG~News and SST-2 and measure principal angles to a reference
subspace estimated from the full training set. Real activations follow the
same qualitative pattern as the synthetic model, with stable estimation at
low $d/N$ and rapid degradation near $d/N\approx 10$ in the evaluated
configurations. The onset of degradation depends on the covariance spectrum
and sampling conditions and is not determined by $d/N$ alone.

\paragraph{Subspace-task alignment.}
A reliably estimated activation subspace is useful only if it also captures
supervised task structure. Let
$\Sigma_{\mathrm{task}}$ denote the between-class scatter of target-token
representations,

\[
  \Sigma_{\mathrm{task}}
  =
  \frac{1}{N}
  \sum_{c=1}^{C}
  n_c(\mu_c-\bar\mu)(\mu_c-\bar\mu)^\top,
\]
where $n_c$ is the number of examples in class $c$, $\mu_c$ is the class mean,
and $\bar\mu$ is the global mean.

\begin{definition}[Subspace-task alignment]
\label{def:alignment}
For a rank-$r$ subspace with orthonormal basis
$U\in\mathbb{R}^{d\times r}$, define
\begin{equation}
\label{eq:rho}
  \rho(U,\Sigma_{\mathrm{task}})
  =
  \frac{\mathrm{Tr}(U^\top \Sigma_{\mathrm{task}} U)}
       {\mathrm{Tr}(\Sigma_{\mathrm{task}})}
  \in [0,1].
\end{equation}
The value $\rho$ measures the fraction of between-class signal captured by the
subspace. It equals one when the column span of $U$ contains the range of
$\Sigma_{\mathrm{task}}$. The metric requires discrete class labels and is not
defined for open-ended generation.
\end{definition}

\paragraph{Why direction affects the first LoRA update.}
Consider a LoRA contribution

\[
  \Delta z
  =
  s B A h(x),
  \qquad
  s=\alpha/r,
\]
where $h(x)\in\mathbb{R}^d$ is the target-token representation. Let
$U\in\mathbb{R}^{d\times r}$ have orthonormal columns and initialize
$A=U^\top$, so that the adapter observes the projected features
$U^\top h(x)$. At standard zero-$B$ initialization, $\Delta z=0$ and
$\nabla_A\mathcal{L}=0$ exactly, while $\nabla_B\mathcal{L}$ can be nonzero.
The first adapter update therefore acts on $B$, and its dependence on the
initial subspace enters through the projected features.

Let

\[
  G_U
  =
  \nabla_B\mathcal{L}(0;U)
\]
denote the initial population gradient. Under the class-conditional mean, cross-covariance, centering, and proportionality assumptions stated
in Appendix~\ref{app:first-step}, its squared Frobenius norm satisfies
\begin{equation}
\label{eq:gradenergy}
  \|G_U\|_F^2
  =
  s^2\gamma\,
  \operatorname{Tr}
  \left(
    U^\top\Sigma_{\mathrm{task}}U
  \right)
\end{equation}
for a task-dependent constant $\gamma>0$. Thus, the first-order loss decrease
from an update to $B$ is controlled by the supervised structure captured by
the row span of $A$. Appendix~\ref{app:first-step} gives the derivation and
states the assumptions explicitly.

If $U$ is sampled uniformly from rank-$r$ subspaces independently of the fixed
task matrix, then its expected captured fraction of between-class signal is
$r/d$. 
For BERT-base with $r=8$, the random-subspace expectation is
$r/d\approx0.010$, while the most aligned task reaches $\rho\approx0.84$ in our measurements, as reported in Appendix~\ref{app:alignment}. Direction-aware
initialization is therefore expected to be useful when $\rho$ substantially
exceeds $r/d$ and the relevant activation structure can be estimated from the
few-shot sample.

The initial span also changes slowly. The gradient with respect to $A$ is
proportional to $B$, which starts at zero, so $A$ moves slowly; in our
classification runs its norm stays within 5\% of its initial value until the
pivot (Appendix~\ref{app:pivot-diagnostics}).

\begin{remark}[Two-condition view]
\label{rem:two-cond}
Direction-aware initialization yields meaningful gains when (i) $N$ is small
relative to the representation dimension, so the adapter row span is sensitive
to its initialization, and (ii) the estimated subspace captures between-class
signal, which requires a spectral gap large enough to control the error in
Equation~\eqref{eq:dk} and a moderate to high $\rho$. In our BERT-base
measurements the advantage is largest above $d/N\approx10$. It shrinks as
either condition weakens, and vanilla LoRA is competitive once data are abundant or the source is
misaligned. There direction adds little, consistent with the magnitude account
of \citet{zhang2025loram}.
\end{remark}

\section{Method}
\label{sec:method}

\begin{figure*}[t]
  \centering
  \includegraphics[width=\textwidth]{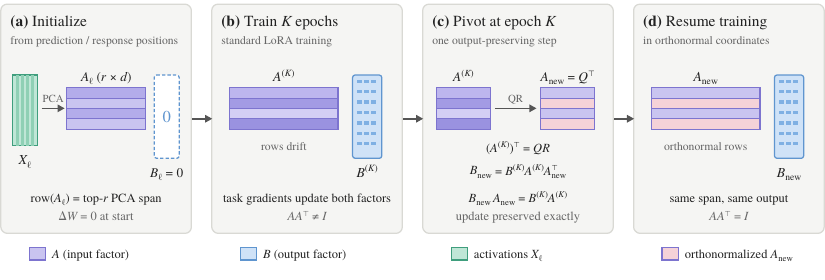}
  \caption{Overview of PivotLoRA. Activations at supervised prediction
  positions initialize the row span of $A_\ell$, with $B_\ell=0$ so that
  $\Delta W_\ell=0$. For classification, the activation source is the \texttt{[CLS]} representation (BERT) or the sequence mean (decoders); for completion-only reasoning, it comprises hidden states that predict supervised response tokens. After
  $K$ epochs, PivotLoRA replaces the learned input basis with an orthonormal
  basis for the same row span and re-encodes $B_\ell$ so that $B_\ell A_\ell$ is preserved.}
  \label{fig:overview}
  \vspace{-0.8em}
\end{figure*}

PivotLoRA operationalizes the two conditions of Section~\ref{sec:theory} by initializing the
LoRA input factor from activations at prediction-relevant positions and
performing one output-preserving orthogonal pivot after early training drift
(Figure~\ref{fig:overview}).

Prediction positions depend on the objective. For BERT classification, we use
the \texttt{[CLS]} representation. Decoder-scale classification, whose label describes the whole sequence, uses the mean of adapter inputs over input positions. Completion-only reasoning uses hidden states that predict supervised response tokens. For the GPT-2 summarization control we compare the sequence mean with response positions (Appendix~\ref{app:generation}).

All methods use the standard LoRA parameterization
\begin{equation}
\label{eq:lora}
  \Delta W = \frac{\alpha}{r}BA,
\end{equation}
where $A\in\mathbb{R}^{r\times d}$ stores input directions as rows and
$B\in\mathbb{R}^{m\times r}$ maps low-rank coordinates to the output space.
Unless stated otherwise, we use rank $r=8$ and scaling $\alpha=16$.

\subsection{Supervised-Position Subspace Initialization}
\label{sec:tt-init}

Activation initializers differ in the positions used to estimate their
subspaces. All-token pooling provides more activation vectors and can estimate
shared structure more reliably, whereas prediction-position pooling targets
the representation directly connected to the supervised objective. Activation-source selection is therefore an alignment--estimation trade-off, and no single source need be optimal. Our primary controlled classification construction uses prediction-position PCA.
A matched ablation shows that all-token and per-projection sources can score higher (Appendix~\ref{app:activation-source}).

Let $\mathcal{D}$ contain $N$ training examples. A frozen forward pass produces
one activation vector $x_{\ell,n}\in\mathbb{R}^{d}$ for each example $n$ and
adapted layer $\ell$. For classification, $x_{\ell,n}$ is the adapter input at the \texttt{[CLS]} position (BERT) or its mean over input positions (decoders). For completion-only reasoning, it
is the pooled response-predictor representation defined in
Section~\ref{sec:response-reasoning}. We form

\[
  X_\ell
  =
  \begin{bmatrix}
    x_{\ell,1}^{\top}\\
    \vdots\\
    x_{\ell,N}^{\top}
  \end{bmatrix}
  \in\mathbb{R}^{N\times d},
  \qquad
  \mu_\ell
  =
  \frac{1}{N}X_\ell^\top\mathbf{1},
  \qquad
  \bar X_\ell
  =
  X_\ell-\mathbf{1}\mu_\ell^\top.
\]
The empirical covariance is
\begin{equation}
\label{eq:pca}
  \widehat{\Sigma}_\ell
  =
  \frac{1}{N}\bar X_\ell^\top\bar X_\ell.
\end{equation}
Let $Q_\ell\in\mathbb{R}^{d\times r}$ contain the leading $r$
eigenvectors of $\widehat{\Sigma}_\ell$, ordered by decreasing eigenvalue:

\[
  \widehat{\Sigma}_\ell Q_\ell
  =
  Q_\ell\Lambda_\ell,
  \qquad
  Q_\ell^\top Q_\ell=I_r.
\]
We initialize
\begin{equation}
\label{eq:pca-init}
  A_\ell^{(0)}=c\,R_\ell^\top Q_\ell^\top,
  \qquad
  B_\ell^{(0)}=\mathbf{0},
  \qquad
  c=\sqrt{5/d},
\end{equation}
where $R_\ell\in\mathbb{R}^{r\times r}$ is a fixed random orthogonal matrix
drawn once per adapter. Because $R_\ell$ is orthogonal, the row span of
$A_\ell^{(0)}$ is exactly the leading PCA subspace; $R_\ell$ only mixes
coordinates within it. The scale gives
$\|A_\ell^{(0)}\|_F=\sqrt{5r/d}$, smaller than the approximately
$\sqrt{r/3}$ of the Kaiming-uniform initialization used by vanilla LoRA
(about $7\times$ smaller for $d=768$ and $17\times$ for $d=4096$). These
ratios hold for the reasoning code. In our classification code, vanilla
LoRA's $A$ is stored as a $d\times r$ tensor, so Kaiming's fan-in is $r$ and
$\|A\|_F\approx\sqrt{d/3}$, about $16$ for BERT and $37$ for Llama; there
PivotLoRA starts about $70\times$ and $370\times$ smaller. On BERT, vanilla
LoRA with the standard initialization is the weaker baseline, exceeding this
version in only 6 of 30 settings, and PivotLoRA exceeds it in 28 of 30
(Appendix~\ref{app:baselines-released}). The
initialization is therefore not norm-matched to vanilla LoRA, and because it
is smaller, PivotLoRA's gains do not come from a larger initialization norm.
Since $B_\ell^{(0)}=\mathbf{0}$, initialization leaves the pretrained model function unchanged. We call this initialization without the pivot TT-PCA (target-token PCA). We prefer PCA to LDA for three reasons. PCA needs no labels, so the same construction applies to completion-only reasoning, where there are no classes. It keeps $\rho$ an independent measure of how much supervised structure an unsupervised subspace captures. LDA also yields at most $C-1\le3$ directions here, fewer than $r=8$. The TT-LDA ablation (Appendix~\ref{app:tt-lda}) has a better average rank than TT-PCA but uses a different learning-rate structure (Appendix~\ref{app:lr-structure}).

In the main classification protocol, the PCA eigenvalues, in decreasing
order, determine an allocation of the learning rate across the $r$ columns of $B$. Because of $R_\ell$, the input direction paired with each column mixes several principal directions. The
average $B$-to-$A$ learning-rate ratio is
$\lambda_{\mathrm{ratio}}=16$, matching the LoRA+ scale
of~\citet{hayou2024loraplus}. This allocation is an optimization heuristic and is not part of the subspace definition.
Classification-specific learning-rate equations appear in Appendix~\ref{app:classification-lr}.

\subsection{Response-Position Initialization for Generative Reasoning}
\label{sec:response-reasoning}

For completion-only fine-tuning, each sequence contains a prompt followed by a
supervised response, with prompt tokens excluded from the language-modeling
loss. Because hidden state $t$ predicts token $t+1$, let
$M_{n,t+1}\in\{0,1\}$ indicate whether token $t+1$ is supervised. The
corresponding predictor positions are

\[
  \mathcal{P}_n
  =
  \left\{t : M_{n,t+1}=1\right\}.
\]
For adapted layer $\ell$, let
$h_{\ell,n,t}\in\mathbb{R}^{d}$ denote the adapter input at predictor position
$t$. PivotLoRA forms one activation vector per training example:
\begin{equation}
\label{eq:response-pooling}
  x_{\ell,n}
  =
  \frac{1}{|\mathcal{P}_n|}
  \sum_{t\in\mathcal{P}_n}h_{\ell,n,t}.
\end{equation}
We then center the resulting example-level activation matrix and compute its
leading $r$ principal directions using Eq.~\eqref{eq:pca}. For the reasoning
experiments, these directions initialize
$A_\ell^{(0)}=\sqrt{5/d}\,Q_\ell^\top$, with
$B_\ell^{(0)}=\mathbf{0}$, the same scale as Eq.~\eqref{eq:pca-init} without the rotation.
Appendix~\ref{app:reasoning-details} gives the prompt templates, masking
rules, generation settings, and remaining training configuration.

\subsection{Output-Preserving Orthogonal Pivot}
\label{sec:pivotlora}

Because $A$ remains trainable, its initially orthogonal rows can become
anisotropic during early optimization. PivotLoRA performs a single
reparameterization after $K$ epochs while preserving both the learned adapter
update and the model function.

Let $A^{(K)}\in\mathbb{R}^{r\times d}$ and
$B^{(K)}\in\mathbb{R}^{m\times r}$ denote the factors at the pivot. Assuming
that $A^{(K)}$ has full row rank, we compute the reduced QR decomposition of
its transpose (in double precision for reasoning and in float32 for
classification):
\begin{equation}
\label{eq:pivot-qr}
  \bigl(A^{(K)}\bigr)^\top=QR,
  \qquad
  Q^\top Q=I_r,
\end{equation}
and define

\[
  A^{\mathrm{new}}=Q^\top.
\]
The rows of $A^{\mathrm{new}}$ are orthonormal and have the same row span as
$A^{(K)}$. We re-encode the output factor as
\begin{equation}
\label{eq:reproj}
  B^{\mathrm{new}}
  =
  B^{(K)}A^{(K)}
  \bigl(A^{\mathrm{new}}\bigr)^\top.
\end{equation}
Because $A^{\mathrm{new}}$ is an orthonormal basis for the row span of
$A^{(K)}$, we have $A^{(K)}=A^{(K)}(A^{\mathrm{new}})^\top A^{\mathrm{new}}$,
and therefore
\begin{equation}
\label{eq:pivot-preservation}
  B^{\mathrm{new}}A^{\mathrm{new}}
  =
  B^{(K)}A^{(K)}.
\end{equation}
The fixed scale $\alpha/r$ in Eq.~\eqref{eq:lora} is unchanged. The pivot
therefore preserves the LoRA update, and hence the model function, up to
floating-point error.

Although the represented update is unchanged, subsequent factorized
optimization need not be invariant to the reparameterization. In particular,
if $G=\nabla_{\Delta W}\mathcal{L}$, then

\[
  \nabla_B\mathcal{L}
  =
  \frac{\alpha}{r}GA^\top.
\]
The geometry of the $B$-factor update therefore depends on the Gram matrix
$AA^\top$. The pivot restores

\[
  A^{\mathrm{new}}
  \bigl(A^{\mathrm{new}}\bigr)^\top
  =
  I_r,
\]
removing coordinate anisotropy from this Gram matrix without replacing the
learned row span. Concretely, a plain gradient step on $B$ changes the update by
$-\eta(\alpha/r)^2GA^\top A$, so before the pivot each direction in the row
span is scaled by the corresponding squared singular value of $A$. After the
pivot, $A^\top A$ is the orthogonal projector onto that span, and every
learned direction receives the same step. The pivot also returns the rows of $A$ to unit norm, which enlarges the effective step on $B$. In classification, the norm of $A$ changes little before the pivot, so the pivot raises it about twelvefold. In reasoning, $A$ grows about sixfold and its Gram matrix has an effective rank of about two or fewer out of eight, and the pivot restores all eight (Appendix~\ref{app:pivot-diagnostics}).

Optimizer-state handling is distinct from the algebraic preservation
guarantee. In the classification implementation, we reset the Adam moments for
$A$ and retain those for $B$. In the reasoning
implementation, we reset the optimizer states for both transformed factors
while leaving the global learning-rate schedule unchanged. In the classification ablation (Appendix~\ref{app:ablation-reset}), the
QR pivot itself accounts for most of PivotLoRA's improvement over TT-PCA, while
the optimizer-state reset produces smaller, task-dependent differences; we therefore regard the reset as an optional stabilizer. The pivot itself takes
under a second, and most of PivotLoRA's 5.8\% wall-time overhead on BERT-base
comes from the per-rank optimizer groups (Appendix~\ref{app:cost}).

For each pivoted adapter we record the relative update-preservation error
and the row-orthogonality error of $A^{\mathrm{new}}$; both stay below
$10^{-6}$ in every run (Appendix~\ref{app:pivot-diagnostics}). Unless stated otherwise, PivotLoRA pivots at $K=3$.

\section{Experiments}
\label{sec:experiments}

We evaluate PivotLoRA on classification and generation tasks with four pretrained models. BERT-base provides the main controlled regime analysis,
decoder-scale classification tests transfer to larger causal models, GPT-2 summarization tests whether a large $d/N$ suffices, and GSM8K and
StrategyQA test the response-position extension for generated-answer
reasoning.

\subsection{Setup}
\label{sec:setup}

\paragraph{Models, tasks, and protocol.}
We evaluate BERT-base~\citep{devlin2019bert} on five GLUE tasks (SST-2, CoLA, MNLI, QNLI, and RTE)~\citep{wang2019glue} and AG~News~\citep{zhang2015agnews} at $N\in\{16,32,64,128,256\}$ total labeled examples, sampled approximately class-balanced, with BoolQ as an auxiliary seventh task for the alignment analysis. We evaluate Llama-2-7B~\citep{touvron2023llama2} and
Llama-3.1-8B~\citep{dubey2024llama3} on SST-2 and AG~News at
$N\in\{64,256\}$, and additionally evaluate Llama-3.1-8B on
BoolQ~\citep{clark2019boolq}. We evaluate
GPT-2~\citep{radford2019language} on XSum~\citep{narayan2018xsum} and
CNN/DailyMail~\citep{hermann2015cnn,nallapati2016cnn} with
$N\in\{64,256\}$ training sequences. Finally, we evaluate completion-only
reasoning with Llama-3.1-8B on GSM8K~\citep{cobbe2021training} and
StrategyQA~\citep{geva2021strategyqa} using $N=128$ training sequences, three
paired seeds, and the complete test split of each dataset.

The primary BERT benchmark contains 30 task--$N$ settings over three seeds.
All methods use rank $r=8$, query and value adapters, AdamW, 15 epochs, and
batch size 16. The shared BERT base learning-rate anchor is
$\eta_A=2\times10^{-4}$. Unless a baseline specifies otherwise,
$\alpha_{\mathrm{LoRA}}=16$. We compare LoRA, DoRA, AdaLoRA, PiSSA, LoRA-GA,
LoRA-One, LoRA+, EVA, LoRAM, and SRLoRA, plus the TT-PCA and TT-LDA ablations.
Method-specific learning-rate allocations, the shared initialization scaling, other implementation settings,
sampling, hardware, and evaluation details appear in
Appendix~\ref{app:implementation}.

For generated-answer reasoning, paired LoRA and PivotLoRA runs share the
sampled examples, base AdamW configuration, learning-rate schedule,
tokenization, completion mask, training budget, decoding procedure, and answer
scoring. Response-position PivotLoRA additionally applies PCA initialization,
one output-preserving pivot after epoch $K=3$, and its prescribed post-pivot
optimizer-state reset. Unlike the classification protocol, neither reasoning
method uses differential or eigenvalue-scaled learning rates.

\subsection{Few-Shot Classification on BERT-Base}
\label{sec:bert-results}

Table~\ref{tab:main} summarizes all six primary tasks, reporting each
method's mean score across tasks at each $N$ (accuracy, or Matthews
correlation for CoLA), its average rank over the 30 task--$N$ settings, and
the number of settings in which it exceeds vanilla LoRA.
Per-task results with standard deviations appear in
Appendix~\ref{app:fullresults}. Under the shared protocol,
no prior initialization method consistently outperforms vanilla LoRA over all
30 settings. TT-PCA surpasses the published baselines on average, supporting activation-based initialization in this regime.

PivotLoRA achieves the best average rank and improves over vanilla LoRA in 29
of 30 primary settings (Appendix~\ref{app:fullrankings}). It is strongest at
$N \leq 64$, with an average rank of 1.39 and the best or tied-best result in 13 of 18 settings. The margins narrow at $N \in \{128,256\}$ as $d/N$ decreases. On AG~News at
$N=64$, for example, PivotLoRA reaches about 89\% of the full-data accuracy,
compared with 68\% for LoRA (Appendix~\ref{app:data-efficiency}).

\begin{table}[!t]
\centering
\begin{minipage}[t]{0.56\textwidth}
\centering
\caption{BERT-base, mean over six tasks and three seeds at each $N$; rank
and ${>}$LoRA over 30 settings. \textbf{Bold}/\underline{underline}: best/second.}
\label{tab:bert-primary-results}
\label{tab:main}
\footnotesize
\setlength{\tabcolsep}{2.2pt}
\begin{tabular}{@{}lccccc@{\hspace{7pt}}cc@{}}
\toprule
& \multicolumn{5}{c}{Labeled examples $N$} & & \\
\cmidrule(lr){2-6}
Method & 16 & 32 & 64 & 128 & 256 & Rank$\downarrow$ & ${>}$LoRA \\
\midrule
LoRA      & .402 & .428 & .488 & .566 & .618 & 5.23 & -- \\
DoRA      & .404 & .423 & .486 & .561 & .614 & 5.83 & 10/30 \\
AdaLoRA   & .379 & .381 & .407 & .433 & .458 & 11.30 & 2/30 \\
PiSSA     & .383 & .395 & .426 & .499 & .581 & 9.87 & 2/30 \\
LoRA-GA   & .383 & .391 & .427 & .505 & .601 & 9.30 & 5/30 \\
LoRA-One  & .379 & .388 & .431 & .523 & .606 & 8.47 & 6/30 \\
LoRA+     & \textbf{.459} & \underline{.481} & .438 & .425 & .415 & 8.43 & 9/30 \\
EVA       & .381 & .391 & .429 & .520 & .606 & 8.47 & 5/30 \\
LoRAM     & .379 & .387 & .434 & .517 & .608 & 8.43 & 6/30 \\
SRLoRA    & .401 & .423 & .472 & .554 & .617 & 6.93 & 4/30 \\
\midrule
TT-PCA    & .415 & .445 & .546 & \underline{.606} & .631 & 3.77 & 24/30 \\
TT-LDA    & .415 & .455 & \underline{.555} & \textbf{.607} & \underline{.634} & \underline{3.13} & \underline{25/30} \\
PivotLoRA & \underline{.451} & \textbf{.500} & \textbf{.575} & \underline{.606} & \textbf{.641} & \textbf{1.83} & \textbf{29/30} \\
\bottomrule
\end{tabular}
\end{minipage}\hfill
\begin{minipage}[t]{0.42\textwidth}
\centering
\caption{Decoder-scale classification at $N=64$, mean over three seeds.}
\label{tab:decoder-main}
\footnotesize
\setlength{\tabcolsep}{2.8pt}
\begin{tabular}{@{}lcccc@{}}
\toprule
& \multicolumn{2}{c}{Llama-2-7B} & \multicolumn{2}{c}{Llama-3.1-8B} \\
\cmidrule(lr){2-3}\cmidrule(lr){4-5}
Method & SST-2 & AG News & SST-2 & AG News \\
\midrule
LoRA      & .708 & .748 & .704 & .759 \\
PiSSA     & .635 & .700 & .622 & .715 \\
LoRAM     & .763 & .742 & .662 & .748 \\
EVA       & .713 & .745 & .643 & .743 \\
\midrule
PivotLoRA & \textbf{.769} & \textbf{.784} & \textbf{.709} & \textbf{.799} \\
\bottomrule
\end{tabular}

\vspace{8pt}
\caption{Llama-3.1-8B reasoning at $N{=}128$: accuracy (\%) over three seeds, $\pm$ one sample SD. McNemar tests are exact and pooled over seeds.}
\label{tab:reasoning}
\footnotesize
\setlength{\tabcolsep}{5pt}
\begin{tabular}{@{}lcc@{}}
\toprule
Method & GSM8K & StrategyQA \\
\midrule
Base (no adapter) & 25.6 & 54.4 \\
LoRA      & 28.7 $\pm$ 0.2 & 64.1 $\pm$ 2.3 \\
PivotLoRA & \textbf{33.2} $\pm$ 2.5 & \textbf{67.7} $\pm$ 1.4 \\
\midrule
Paired gain & $+4.5 \pm 2.3$ & $+3.6 \pm 3.1$ \\
Seeds improved & 3/3 & 2/3 (1 tie) \\
McNemar $p$ & $<10^{-6}$ & 0.004 \\
\bottomrule
\end{tabular}
\end{minipage}
\vspace{-0.6em}
\end{table}

\paragraph{Sensitivity and full fine-tuning.}
At $N=64$, a representative learning-rate sweep narrows some shared-protocol
margins, but PivotLoRA remains strongest on average over AG~News, SST-2, and
CoLA among the swept methods, with 0.667 versus 0.655 for TT-PCA, 0.636 for
LoRA+, and 0.622 for LoRA
(Appendices~\ref{app:lr-sweep} and~\ref{app:loraplus}). At the same $N$,
PivotLoRA exceeds the mean full-fine-tuning score on all six primary tasks while training approximately 0.30M adapter parameters
(Appendix~\ref{app:fullft}).

\paragraph{Direction versus learning-rate allocation.}
At $N=64$, a random input basis drawn as in vanilla LoRA and given the same eigenvalue-scaled
learning rates reaches 0.444, 0.571, and 0.028 on AG~News, SST-2, and CoLA,
versus 0.751, 0.769, and 0.268 for TT-PCA, while TT-PCA with uniform or
permuted rates is essentially unchanged (Appendix~\ref{app:direction-lr-controls}).
The PCA directions, not the rate allocation, carry the gain.

\paragraph{Contribution of the pivot.}
Applied to the same TT-PCA initialization without the optimizer-state reset,
the pivot alone improves the score in 12 of 15 task--$N$ settings, by $0.037$
on average and by up to $0.128$ on AG~News at $N=32$
(Appendix~\ref{app:ablation-reset}). Each tested pivot epoch, $K\in\{1,2,3,5,7,10\}$, outperforms LoRA at every $N\le64$ on SST-2, CoLA, and AG~News, so the default $K=3$ is not a finely tuned choice (Appendix~\ref{app:ablation-pivot}). The gain also does not depend on how the $B$ learning rate is handled after the pivot, since leaving it unchanged gives $+0.036$ over TT-PCA (Appendix~\ref{app:post-pivot-lr}).

\paragraph{Robustness.}
A two-sided sign test on the 29-of-30 record gives $p=5.8\times10^{-8}$, and
five-seed reruns of seven representative settings yield bootstrap 95\%
intervals for the paired gain over LoRA that exclude zero in every case,
with mean gains from $+0.032$ on QNLI to $+0.231$ on AG~News
(Appendices~\ref{app:significance} and~\ref{app:5seed}). Restoring LoRAM's
intended initialization magnitude does not close its gap to LoRA
(Appendix~\ref{app:implementation}), so initialization magnitude alone does not explain the gains.

\subsection{Decoder-Scale Classification}
\label{sec:decoder-main}

The activation-subspace pattern also appears in decoder-scale classification
(Table~\ref{tab:decoder-main}). At $N=64$, PivotLoRA ranks first on both
backbones, while EVA does not close the aggregate gap under the same per-layer
rank budget. At $N=256$, PivotLoRA trails LoRA in three of the four SST-2 and AG~News settings, whereas on BoolQ with Llama-3.1-8B PivotLoRA leads LoRA at both $N=64$ and
$N=256$ (.602 versus .554 and .618 versus .576). Complete results appear in Appendix~\ref{app:decoder}. Because the decoder activation source is a sequence mean, these gains suggest that a coarse source suffices when the label describes the whole sequence. At $N=256$, the all-token EVA initializer becomes competitive, consistent with the alignment--estimation trade-off, since a larger sample favors the source with more activation vectors.

\subsection{Generated-Answer Reasoning}
\label{sec:reasoning-results}

\begin{figure}[!t]
\centering
\begin{minipage}[c]{0.49\textwidth}
\centering
\includegraphics[width=\linewidth]{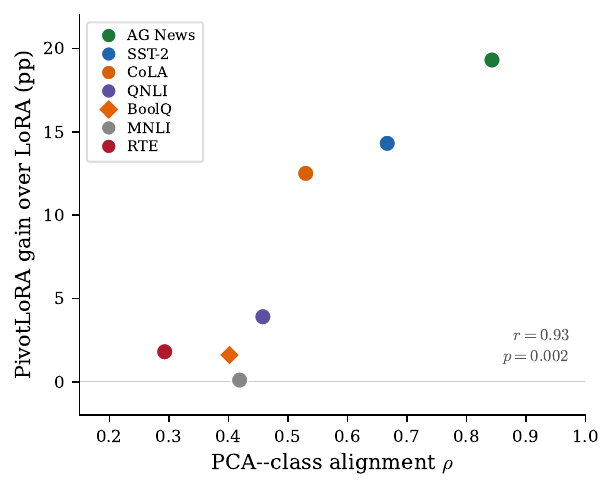}
\vspace{-1.2em}
\captionof{figure}{Alignment $\rho$ versus PivotLoRA's gain over LoRA across
seven BERT tasks at $N=64$ (CoLA in Matthews correlation points).}
\label{fig:rho}
\end{minipage}\hfill
\begin{minipage}[c]{0.48\textwidth}
\centering
\captionof{table}{Regime map. Mean gain of PivotLoRA over LoRA in points of
each task's primary metric, and settings won (details in Table~\ref{tab:regime-detail}). $d/N$ is comparable within a backbone.}
\label{tab:regime}
\footnotesize
\setlength{\tabcolsep}{3pt}
\begin{tabular}{@{}lcccc@{}}
\toprule
Setting & $d/N$ & Source & Gain & Wins \\
\midrule
BERT, $N\le64$      & 12--48 & \texttt{[CLS]}    & $+6.9$ & 18/18 \\
BERT, $N\ge128$     & 3--6   & \texttt{[CLS]}    & $+3.2$ & 11/12 \\
BERT, full data     & ${<}0.4$ & \texttt{[CLS]}  & $+0.4$ & 3/6 \\
\midrule
Llama-2/3.1, $N=64$ & 64 & seq.\ mean    & $+3.8$ & 5/5 \\
Llama-2/3.1, $N=256$ & 16 & seq.\ mean    & $+0.4$ & 2/5 \\
Llama-3.1, reasoning    & 32     & response  & $+4.1$ & 2/2 \\
\midrule
GPT-2, summarization & 3--12 & response & $+0.3$ & 3/4 \\
\bottomrule
\end{tabular}
\end{minipage}
\vspace{-0.8em}
\end{figure}

Table~\ref{tab:reasoning} reports completion-only fine-tuning with
Llama-3.1-8B on GSM8K (mathematical word problems with worked solutions) and
StrategyQA (yes/no questions supervised with supporting facts). Within each
dataset and seed, LoRA and PivotLoRA share the same $N=128$ training examples
and are evaluated on the complete test split (1,319 and 687 questions).
PivotLoRA improves on LoRA on both datasets, at every seed on GSM8K, and an exact item-level McNemar test pooled over seeds rejects equal accuracy on both. Pooled over the three seeds, PivotLoRA alone answers 655 GSM8K test items correctly against 475 for LoRA alone, and 350 against 276 on StrategyQA. Both methods follow the
trained answer format in at least 95\% of generations, so formatting does not explain the gains. These comparisons evaluate the full procedure (Appendix~\ref{app:reasoning-details}). In these runs, early training concentrates each input factor on about two or fewer of its eight directions before the pivot, and the pivot restores an isotropic basis for the same span (Appendix~\ref{app:pivot-diagnostics}).

\subsection{Regime Summary and Alignment Diagnostic}
\label{sec:diagnostic}

Table~\ref{tab:regime} organizes every evaluated setting by $d/N$ and activation source, following Remark~\ref{rem:two-cond}. Within each backbone the gain shrinks as $d/N$
falls, and with full training sets PivotLoRA, LoRA, and PiSSA perform
comparably (Appendix~\ref{app:fulldata}). On GPT-2 summarization, PivotLoRA is competitive with LoRA but not decisively better under the single learning rate used for reasoning ($+0.2$ to $+0.3$ ROUGE-L points), and it trails under the classification learning-rate recipe whichever positions it pools (Appendix~\ref{app:generation}). Because the same response-position source helps reasoning, the difference may reflect the backbone, the adapted modules, or how much loss-relevant structure the leading directions capture in summarization. A large $d/N$ is therefore not sufficient by itself.

Across the six primary BERT tasks plus auxiliary BoolQ, the classification
alignment coefficient $\rho$ at $N=64$ is strongly associated with
PivotLoRA's shared-protocol gain over vanilla LoRA, with Pearson $r=0.93$ and
exact permutation $p=0.002$ as shown in Figure~\ref{fig:rho}. This is consistent with Equation~\eqref{eq:gradenergy}, in which $\rho$ is the share of between-class signal available to the first update of $B$. Spearman correlation is 0.86, leave-one-task-out Pearson correlations stay within $[0.87,0.96]$, and a cross-split estimate preserves the task ordering (Appendices~\ref{app:alignment} and~\ref{app:rho-robustness}). Both $d/N$ and $\rho$ are available before training. When $d/N$ is large and $\rho$ stays high across resampled subsets, direction-aware initialization is likely to help, and when either is small, a well-tuned vanilla LoRA is a strong default.

\FloatBarrier

\raggedbottom
\section{Conclusion}

Initialization direction is not uniformly consequential for LoRA. It matters
when few labeled examples leave the adapter's row span sensitive to its
starting point and the estimated activation subspace carries supervised
structure, and it fades once data are abundant or the activation source is
poorly aligned with the objective. An alignment measure that does not need
class labels, such as one built from loss gradients at initialization, would
extend the diagnostic to generative tasks, and calibrating it across backbones
would turn the regime account into a decision rule. Because the pivot applies
to any full-row-rank starting basis, it could also test whether these regime
boundaries hold for weight- or gradient-based initializers.

\begin{ack}
\paragraph{Funding and support.}
We thank the University of Delaware for providing computational resources.
Peng Zhao's work was partially supported by a research gift from Microsoft
Research.

\paragraph{Competing interests.}
The authors declare no competing interests.
\end{ack}

\bibliographystyle{plainnat}
\bibliography{references}

\newpage
\appendix
\clearpage

\section{Extended Related Work}
\label{app:related}

\paragraph{LoRA initialization.}
Vanilla LoRA~\citep{hu2022lora} initializes $A$ with Kaiming-uniform noise and
$B=0$, so that $\Delta W=0$ at the start of training. Subsequent work has
explored several sources of initialization signal.

\emph{Weight-based methods} derive the adapter subspace from the pretrained
weight matrix $W_0$. PiSSA~\citep{meng2024pissa} initializes $A$ and $B$ from
the top-$r$ singular components of $W_0$ and freezes the residual
$W^{\mathrm{res}}=W_0-W_0^{(r)}$. MiLoRA~\citep{wang2025milora} takes the
complementary approach and initializes from minor singular components to
preserve the dominant directions of $W_0$. Both methods are task-agnostic and derive their subspaces from pretrained weight geometry alone. Because our benchmark already includes PiSSA as a representative
weight-spectral baseline, we do not evaluate MiLoRA separately.

\emph{Gradient-based methods} use the downstream task loss.
LoRA-GA~\citep{wang2024loraga} computes a forward--backward pass and initializes
from the SVD of $\nabla_W\mathcal L$. LoRA-One~\citep{zhang2025loraone}
uses balanced singular-value scaling between $A$ and $B$.
LoRA-DA~\citep{zhang2026lorada} augments the gradient with a diagonal Fisher
estimate and solves a quadratic program for initialization.
Appendix~\ref{app:lora-da} reports our LoRA-DA implementation under the
few-shot protocol used in this work.

\emph{Activation-based methods} estimate a subspace from the input
distribution. EVA~\citep{paischer2025eva} collects activations from all token
positions in a calibration minibatch, computes an SVD of their covariance,
and initializes $A$ from the resulting right singular vectors. The original
method also permits adaptive redistribution of rank across layers. In our
controlled comparison, we disable redistribution and use uniform rank
$r=8$ at every adapted layer. CorDA~\citep{yang2024corda} modulates the
weight SVD by an activation covariance, thereby orienting the decomposition
toward downstream input statistics. CorDA requires covariance inversion,
which was numerically unstable in our smallest-$N$ setting, so it is not
included in the primary comparison.

Activation-based initializers also differ in which token positions supply
the activation statistics. All-token pooling, as used by EVA, provides more
activation observations, whereas PivotLoRA's prediction-position pooling
targets representations associated with the supervised objective. This
distinction introduces an alignment--estimation trade-off, with no universal advantage for either source.

LoRAM~\citep{zhang2025loram} provides a magnitude-based
perspective. It argues that gains from pretrained weight-spectral
initializers can often be explained by update-magnitude amplification, not by the semantic content of the selected singular directions.
LoRAM approximates these gains using magnitude-scaled orthogonal bases
without storing spectral components. Our setting is complementary.
PivotLoRA studies directions estimated from downstream activation
covariance and includes controls that separate prediction-position
direction from eigenvalue-scaled learning rates.

\paragraph{Dynamic adapter methods.}
SRLoRA~\citep{yang2025srlora} periodically fuses low-importance rank-one pairs
into the frozen weights and reinitializes from unused SVD directions.
Flora~\citep{hao2024flora} resamples random projections at fixed intervals.
PivotLoRA instead performs one output-preserving orthogonal
reparameterization. It replaces the learned input basis with an orthonormal
basis for the same row span and transforms $B$ so that the represented update
is preserved up to numerical precision. The pivot is therefore a coordinate change, unlike periodic rank replacement or random resampling.

\paragraph{Structural and optimization variants.}
DoRA~\citep{liu2024dora} decomposes the adapted weight into magnitude and
direction components. AdaLoRA~\citep{zhang2023adalora} parameterizes adapters
as SVD triplets and prunes singular values during training.
LoRA+~\citep{hayou2024loraplus} retains the standard factorization but uses
different learning rates for the two factors, typically with
$\eta_B/\eta_A\approx16$. These methods alter the adapter parameterization,
rank allocation, or factor-wise optimization rule. PivotLoRA retains the
standard low-rank update and instead changes how its input-side coordinates are
initialized and, once during training, orthogonally reparameterized.

Learning-rate selection can materially affect comparisons among LoRA methods.
\citet{lee2026learningrate} systematically re-evaluate vanilla LoRA
and nine variants over learning rate, batch size, rank, and training duration.
They find that different methods favor different learning-rate ranges and that,
after broad tuning, their peak performance is often similar. This result
cautions against interpreting improvements obtained under one shared
configuration as evidence of an intrinsic methodological advantage. It also motivates our method-specific learning-rate sensitivity analysis (Appendix~\ref{app:lr-sweep}), which tests whether the principal comparisons persist over a prespecified range under our few-shot protocol.

\paragraph{Subspace estimation in high dimensions.}
The Davis--Kahan $\sin\Theta$ theorem~\citep{davis1970rotation,yu2015useful}
bounds the angle between estimated and population eigenspaces under spectral
perturbation. Under spiked covariance models~\citep{johnstone2001distribution},
PCA recovery depends on sample size, effective dimension, and spectral
separation. In proportional asymptotic regimes, eigenvectors can also undergo
a phase transition near the BBP threshold~\citep{baik2005phase}. We use these
results as regime-level motivation for dependence on
$d_{\mathrm{eff}}/N$, not as a complete probabilistic model of transformer
fine-tuning.

\FloatBarrier

\section{Additional Theory}
\label{app:theory}

\subsection{Empirical Variance Concentration in \texttt{[CLS]} Activations}
\label{app:spiked-validation}

Section~\ref{sec:theory} models prediction-position activations through the
spiked covariance form

\[
  \Sigma
  =
  \sum_{i=1}^{r^\ast}
  (\lambda_i-\sigma^2)v_i v_i^\top
  +
  \sigma^2 I,
\]
where $\lambda_i>\sigma^2$, the vectors
$v_1,\ldots,v_{r^\ast}$ are orthonormal signal directions, and
$\sigma^2 I$ is an isotropic noise floor. Under this parameterization, the
population eigenvalue along $v_i$ is $\lambda_i$, and each remaining
population eigenvalue is $\sigma^2$.

The model is idealized but captures the possibility that substantial
activation variance is concentrated in a low-dimensional subspace. We examine
this premise using frozen BERT-base \texttt{[CLS]} representations.

\begin{figure}[htbp]
  \centering
  \includegraphics[width=0.95\columnwidth]
    {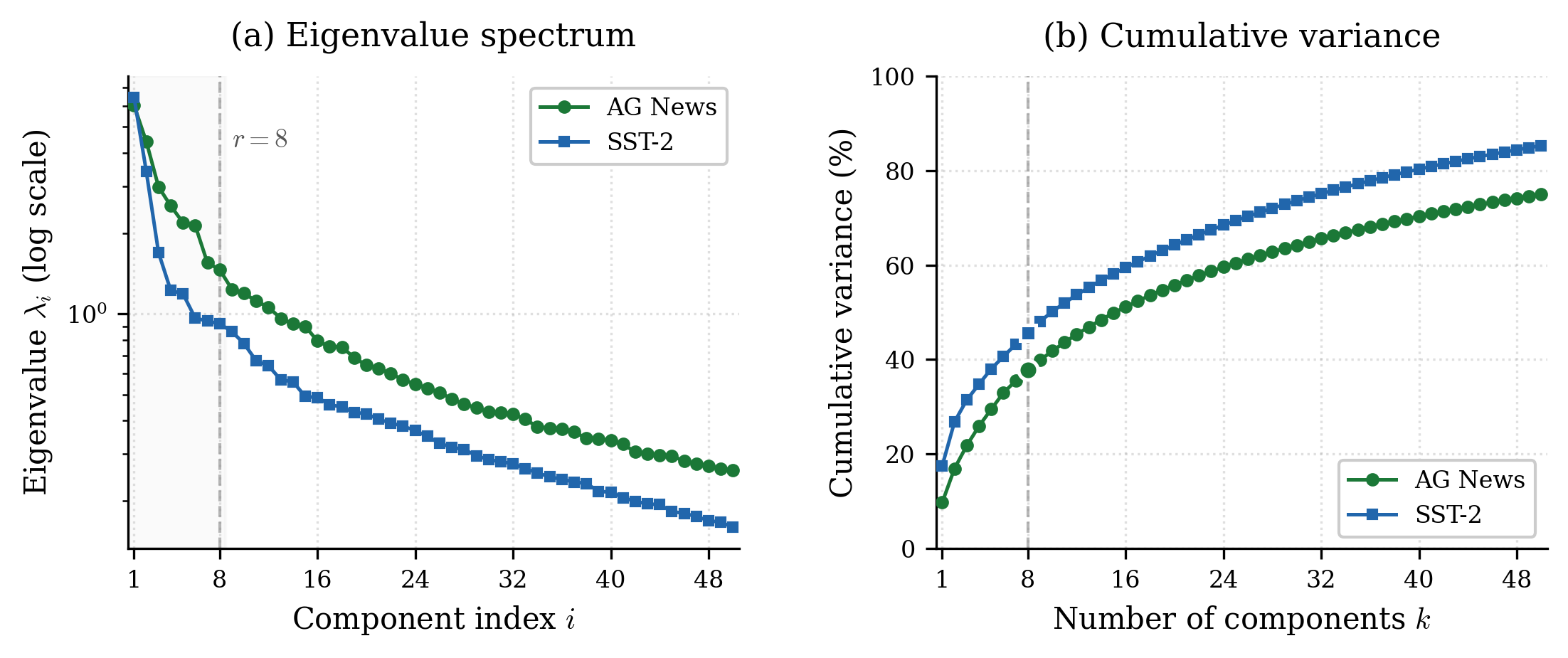}
  \caption{Eigenvalue spectrum and cumulative variance of frozen BERT-base
  \texttt{[CLS]} activations at $N=64$. The top $r=8$ components capture
  38\% of the total variance for AG~News and 46\% for SST-2. The spectra decay continuously, with no sharp finite-rank cutoff.}
  \label{fig:app-scree}
\end{figure}

We run the frozen encoder on $N=64$ total labeled examples from AG~News and
SST-2, sampled approximately class-balanced. We extract the final-layer
\texttt{[CLS]} representation, construct

\[
  \bar X
  =
  X-\mathbf 1\mu^\top
  \in\mathbb R^{N\times768},
  \qquad
  \mu=N^{-1}X^\top\mathbf 1,
\]
and compute

\[
  \widehat\Sigma
  =
  N^{-1}\bar X^\top\bar X.
\]
Figure~\ref{fig:app-scree} reports the leading eigenvalues and cumulative
explained variance.

\FloatBarrier

The top eight components use approximately one percent of the 768-dimensional
representation space while capturing 38--46\% of total variance. Because $N=64$, the sample covariance has rank at most 63, so this share is measured over at most 63 nonzero sample eigenvalues. This indicates substantial low-dimensional concentration without an exact finite-rank cutoff, so we use the spiked model and the perturbation calculation below as regime-level guidance.

\subsection{Derivation of Equation~\eqref{eq:dk}}
\label{app:proof-dk}

Let $X_1,\ldots,X_N\in\mathbb R^d$ be independent, centered,
sub-Gaussian activation vectors with covariance $\Sigma$. Let
$\widehat\Sigma$ be their sample covariance, let $V^\ast$ contain the leading
$r^\ast$ population eigenvectors, and let $\widehat V$ contain the
corresponding sample eigenvectors. We take the adapter rank equal to the number of spikes, $r=r^\ast$. For $r\neq r^\ast$ the gap is $\lambda_r-\lambda_{r+1}$. Under the spiked model, the eigengap is

\[
  \delta
  =
  \lambda_{r^\ast}-\sigma^2
  >0.
\]

An operator-norm Davis--Kahan bound gives

\[
  \left\|
    \sin\Theta(\widehat V,V^\ast)
  \right\|_{\mathrm{op}}
  \lesssim
  \frac{
    \|\widehat\Sigma-\Sigma\|_{\mathrm{op}}
  }{\delta}.
\]
Effective-rank covariance concentration gives

\[
  \mathbb E
  \left[
    \|\widehat\Sigma-\Sigma\|_{\mathrm{op}}
  \right]
  \lesssim
  \|\Sigma\|_{\mathrm{op}}
  \left(
    \sqrt{\frac{d_{\mathrm{eff}}}{N}}
    +
    \frac{d_{\mathrm{eff}}}{N}
  \right),
\]
where

\[
  d_{\mathrm{eff}}
  =
  \frac{\operatorname{Tr}(\Sigma)}
       {\|\Sigma\|_{\mathrm{op}}}.
\]

For a principal-angle matrix of rank at most $r^\ast$,

\[
  \|\sin\Theta\|_F
  \leq
  \sqrt{r^\ast}\,
  \|\sin\Theta\|_{\mathrm{op}},
  \qquad
  \|\sin\Theta\|_F
  \leq
  \sqrt{r^\ast}.
\]
Combining these inequalities yields

\[
  \mathbb E
  \left[
    \|\sin\Theta(\widehat V,V^\ast)\|_F
  \right]
  \lesssim
  \min
  \left\{
    \sqrt{r^\ast},
    \sqrt{r^\ast}\,
    \frac{\|\Sigma\|_{\mathrm{op}}}{\delta}
    \left(
      \sqrt{\frac{d_{\mathrm{eff}}}{N}}
      +
      \frac{d_{\mathrm{eff}}}{N}
    \right)
  \right\},
\]
which is Equation~\eqref{eq:dk}.

The bound provides stronger control when $N$ is large relative to
$d_{\mathrm{eff}}$ and when the eigengap $\delta$ is large. If either
condition deteriorates, the bound becomes weaker, which is consistent with
greater finite-sample sensitivity of the estimated subspace. Transformer activations need not be independent or exactly sub-Gaussian, and their covariance need not follow a literal spiked model, so the calculation is not a finite-sample theorem for transformer fine-tuning.

The PCA covariance uses all $N$ examples. By contrast, the alignment
diagnostic depends on class means estimated from approximately $N/C$ examples
per class. We therefore express PCA estimation through
$d_{\mathrm{eff}}/N$ while recognizing the additional uncertainty in
class-conditional quantities.

\subsection{First-Step LoRA Signal and Task Alignment}
\label{app:first-step}

This section derives the local first-step relationship used in
Section~\ref{sec:theory}. Consider

\[
  \Delta z
  =
  s\,B U^\top h(x),
  \qquad
  s=\alpha/r,
\]
where $U\in\mathbb R^{d\times r}$ has orthonormal columns and
$B\in\mathbb R^{m\times r}$. This corresponds to $A=U^\top$.

Define

\[
  q(x,y)
  =
  \left.
  \nabla_{\Delta z}
  \ell\!\left(z_0(x)+\Delta z,y\right)
  \right|_{\Delta z=0}
\]
and

\[
  \mathcal L(B;U)
  =
  \mathbb E_{(x,y)}
  \left[
    \ell\!\left(
      z_0(x)+sBU^\top h(x),y
    \right)
  \right].
\]

\begin{proposition}[First-step signal in a fixed LoRA subspace]
\label{prop:first-step}
Assume that $\mathcal L(B;U)$ is twice differentiable near $B=0$ with a
locally bounded Hessian. Let

\[
  G_U
  =
  \nabla_B\mathcal L(0;U).
\]
A gradient step $B_1=-\eta G_U$ from $B_0=0$ satisfies

\[
  \mathcal L(B_1;U)-\mathcal L(0;U)
  =
  -\eta\|G_U\|_F^2
  +
  O\!\left(\eta^2\|G_U\|_F^2\right).
\]

Suppose further that, conditional on class $y=c$,

\[
  h(x)
  =
  \mu_c+\varepsilon,
  \qquad
  q(x,y)
  =
  g_c+\xi,
\]
where

\[
  \mathbb E[\varepsilon\mid y=c]=0,
  \qquad
  \mathbb E[\xi\mid y=c]=0,
\]
and the class-conditional cross-covariance vanishes:

\[
  \mathbb E
  \left[
    \xi\varepsilon^\top
    \mid y=c
  \right]
  =
  0.
\]
Let the class priors be $\pi_c$, define

\[
  \bar\mu
  =
  \sum_{c=1}^{C}\pi_c\mu_c,
\]
and assume the class-gradient codes are centered:

\[
  \sum_{c=1}^{C}\pi_c g_c=0.
\]
Then

\[
  \|G_U\|_F^2
  =
  s^2
  \operatorname{Tr}
  \left(
    U^\top H_{\mathrm{grad}}U
  \right),
\]
where

\[
  H_{\mathrm{grad}}
  =
  \sum_{c,c'=1}^{C}
  \pi_c\pi_{c'}
  \langle g_c,g_{c'}\rangle
  (\mu_c-\bar\mu)(\mu_{c'}-\bar\mu)^\top.
\]
If the gradient-weighted class geometry satisfies

\[
  H_{\mathrm{grad}}
  =
  \gamma\Sigma_{\mathrm{task}}
\]
for some task-dependent $\gamma>0$, then

\[
  \|G_U\|_F^2
  =
  s^2\gamma
  \operatorname{Tr}
  \left(
    U^\top\Sigma_{\mathrm{task}}U
  \right).
\]
\end{proposition}

\begin{proof}
Differentiating through the adapter at $B=0$ gives

\[
  G_U
  =
  s\,
  \mathbb E
  \left[
    q(x,y)(U^\top h(x))^\top
  \right].
\]
Conditioning on $y=c$ and expanding $q=g_c+\xi$ and
$h=\mu_c+\varepsilon$ gives

\[
  \mathbb E[q h^\top\mid y=c]
  =
  g_c\mu_c^\top,
\]
because the conditional residual means and cross-covariance vanish. Hence

\[
  G_U
  =
  s\sum_{c=1}^{C}
  \pi_c g_c(U^\top\mu_c)^\top.
\]
The centering condition on the class-gradient codes permits subtraction of
the global representation mean:

\[
  G_U
  =
  s\sum_{c=1}^{C}
  \pi_c g_c
  \left(
    U^\top(\mu_c-\bar\mu)
  \right)^\top.
\]
Expanding the squared Frobenius norm and rewriting it as a trace yields

\[
  \|G_U\|_F^2
  =
  s^2
  \operatorname{Tr}
  \left(
    U^\top H_{\mathrm{grad}}U
  \right).
\]
The special case
$H_{\mathrm{grad}}=\gamma\Sigma_{\mathrm{task}}$
gives the final equality.

For the loss-decrease statement, a Taylor expansion at $B=0$ gives

\[
  \mathcal L(-\eta G_U;U)
  =
  \mathcal L(0;U)
  -
  \eta\|G_U\|_F^2
  +
  O\!\left(\eta^2\|G_U\|_F^2\right).
\]
\end{proof}

\paragraph{Interpretation.}
The most general quantity in the proposition is
$H_{\mathrm{grad}}$, not the unweighted between-class scatter. The reduction
to $\Sigma_{\mathrm{task}}$ requires the stated proportionality assumption.
The proposition is therefore a local explanatory calculation and not a convergence theorem. It takes $A=U^\top$. For the implemented initialization $A=cR^\top Q^\top$ of Eq.~\eqref{eq:pca-init}, the rotation $R$ cancels and $\|G\|_F^2$ gains a factor $c^2$, so its dependence on $\operatorname{Tr}(Q^\top\Sigma_{\mathrm{task}}Q)$ is unchanged. The proposition uses population priors $\pi_c$, and the empirical $\Sigma_{\mathrm{task}}$ of Section~\ref{sec:theory} replaces them with $n_c/N$.

If $U$ is sampled uniformly from the Grassmannian of rank-$r$ subspaces and is
independent of a fixed positive-semidefinite task matrix, then

\[
  \mathbb E_U
  \left[
    \frac{
      \operatorname{Tr}(U^\top\Sigma_{\mathrm{task}}U)
    }{
      \operatorname{Tr}(\Sigma_{\mathrm{task}})
    }
  \right]
  =
  \frac{r}{d}.
\]
For BERT-base with $r=8$ and $d=768$, this expectation is approximately
$0.010$. The largest measured target-token alignment is approximately
$\rho=0.84$, as reported in Appendix~\ref{app:alignment}.

\subsection{Output Preservation Under the Orthogonal Pivot}
\label{app:proof-pivot}

\begin{proposition}[Exact preservation under orthogonal reparameterization]
\label{prop:preservation}
Let $A \in \mathbb{R}^{r \times d}$ have full row rank and let $B \in \mathbb{R}^{m \times r}$. Let $A^{\text{new}} \in \mathbb{R}^{r \times d}$ be an orthonormal row basis for $\text{row}(A)$, so that
\begin{equation*}
    A^{\text{new}}(A^{\text{new}})^\top = I_r
\end{equation*}
and
\begin{equation*}
    \text{row}(A^{\text{new}}) = \text{row}(A).
\end{equation*}
Define
\begin{equation*}
    B^{\text{new}} = BA(A^{\text{new}})^\top.
\end{equation*}
Then
\begin{equation*}
    B^{\text{new}}A^{\text{new}} = BA.
\end{equation*}
Consequently, the LoRA update $\Delta W = (\alpha/r)BA$ is preserved exactly in exact arithmetic.
\end{proposition}

\begin{proof}
The matrix

\[
  P
  =
  (A^{\mathrm{new}})^\top A^{\mathrm{new}}
\]
is the orthogonal projector onto $\operatorname{row}(A)$. Every row of $A$
lies in that subspace, so $AP=A$. Therefore

\[
  B^{\mathrm{new}}A^{\mathrm{new}}
  =
  BA(A^{\mathrm{new}})^\top A^{\mathrm{new}}
  =
  BAP
  =
  BA.
\]
\end{proof}

Because the construction depends only on the current full-row-rank adapter factors, the pivot can follow any initializer, whether PCA-, weight-, activation-, or gradient-based. In finite precision, the equality is affected by QR and matrix-multiplication error. Appendix~\ref{app:pivot-diagnostics} reports these errors for the classification and reasoning runs.

For $G=\nabla_{\Delta W}\mathcal L$ and, up to the factor $\alpha/r$, $\Delta W=BA$,

\[
  \nabla_B\mathcal L=GA^\top,
  \qquad
  \nabla_A\mathcal L=B^\top G.
\]
The $B$-gradient geometry therefore depends on $AA^\top$. The pivot replaces
the current row coordinates with an orthonormal basis, making
$AA^\top=I_r$ while preserving $BA$. Adam and finite-step SGD are not
generally invariant to this factor reparameterization, so subsequent
optimization can change even though the adapter output is unchanged at the
pivot.

\FloatBarrier

\section{Experimental Details}
\label{app:implementation}

Code, configurations, sampled-example manifests, result summaries, selected run artifacts, and available prediction artifacts are released at \url{https://github.com/Nischal1011/pivotlora}.

\paragraph{Software and hardware.}
The BERT-base and GPT-2 experiments were run on a single NVIDIA T4 with
16\,GB of memory, while decoder-scale classification used a single NVIDIA L4
with 24\,GB. Generated-answer reasoning used one NVIDIA GeForce RTX~5090 with 31.8\,GB of memory, PyTorch \texttt{2.9.0+cu128}, and CUDA~12.8, with the base model in bf16 and the adapter factors in float32.

\paragraph{Reproducibility.}
The classification implementation enables deterministic PyTorch and cuDNN
settings. Main classification results use seeds 42, 43, and 44 with
approximately class-balanced sampling unless otherwise noted. Deterministic
configuration reduces avoidable variation but does not guarantee bitwise
identity across different hardware, driver, and library versions.

\paragraph{Asset licenses.}
BERT-base is released under Apache-2.0, GPT-2 under a modified MIT license,
Llama-2-7B under the Llama~2 Community License, and Llama-3.1-8B under the
Llama~3.1 Community License; we accessed the Llama checkpoints after accepting
these terms. GSM8K and StrategyQA are released under MIT, BoolQ under
CC~BY-SA~3.0, QNLI (derived from SQuAD~\citep{rajpurkar2016squad}) under CC~BY-SA~4.0, MNLI~\citep{williams2018mnli} under the
OANC terms with CC~BY-SA portions, and CNN/DailyMail under Apache-2.0 as
distributed on Hugging Face. SST-2~\citep{socher2013sst}, CoLA~\citep{warstadt2019cola}, RTE~\citep{dagan2006rte}, AG~News, and XSum are
distributed for research use without an explicit license. All assets are used
only for research evaluation, and no model weights or dataset contents are
redistributed with our code.

\paragraph{Adapter targets.}
For BERT-base, we adapt
\texttt{attention.self.query} and
\texttt{attention.self.value}, giving 24 adapters. For Llama-2-7B and
Llama-3.1-8B classification, we adapt
\texttt{self\_attn.q\_proj} and
\texttt{self\_attn.v\_proj}, giving 64 adapters. For GPT-2, we adapt
\texttt{attn.c\_proj} and \texttt{mlp.c\_fc}.

\paragraph{BERT classification details.}
The task-specific linear classification head is trained jointly with the
adapter parameters. Its initialization is matched across methods and seeds.
The reported adapter parameter count excludes the classification head.
Few-shot examples are sampled from the training split. Evaluation uses the
validation split when available, the matched validation split for MNLI, and
the test split for AG~News in the dataset distribution used by our
implementation.

\paragraph{Decoder-scale classification.}
Decoder-scale experiments use sequence-classification models with a
task-specific head trained jointly with the adapters. Sentence-pair fields are
concatenated before tokenization. The activation source is the mean of each adapter's inputs over all positions of the padded input sequence, computed in one frozen forward pass. EVA provides an activation-based comparison, although it differs from PivotLoRA in components beyond activation pooling.

\paragraph{Few-shot sampling convention.}
The BERT columns labeled 16, 32, 64, 128, and 256 denote the total number of
labeled examples $N$, sampled approximately class-balanced. Thus, a
$C$-class task contains approximately $N/C$ examples per class. The PCA
covariance is estimated from all $N$ examples.

\paragraph{Baseline configurations.}
PiSSA~\citep{meng2024pissa} uses the top-$r$ SVD of $W_0$ with a frozen
residual and $\alpha_{\mathrm{LoRA}}=r$.
LoRAM~\citep{zhang2025loram} estimates $\sigma_{\max}$ with 10 power
iterations, constructs a fixed-seed random orthogonal basis, and scales it by
$\log(1+\sigma_{\max})$.
LoRA-GA~\citep{wang2024loraga} uses an SVD of a one-step gradient with
residual subtraction.
LoRA-One~\citep{zhang2025loraone} uses balanced SVD scaling without its
optional Scaled-Adam preconditioner.
EVA~\citep{paischer2025eva} uses all-token activation SVD. We disable adaptive
rank redistribution and set $\rho_{\mathrm{EVA}}=1.0$, so each adapted layer
uses rank $r=8$.
LoRA+~\citep{hayou2024loraplus} uses Kaiming initialization and
$\eta_B=16\eta_A$.
DoRA~\citep{liu2024dora} uses its magnitude-direction parameterization.
AdaLoRA~\citep{zhang2023adalora} uses SVD triplets with rank pruning.
SRLoRA~\citep{yang2025srlora} fuses rank-one pairs every
$\max(5,\lfloor T/10\rfloor)$ optimizer steps, where $T$ is the total number of
training steps.
The LoRA-DA implementation uses diagonal Fisher estimation with residual
subtraction.

\paragraph{Baseline fidelity.}
In our classification implementation, every supplied initialization of $A$,
including those of PiSSA, LoRA-GA, LoRA-One, LoRAM, and EVA, passes through the
same LoRA layer, which multiplies it by $\sqrt{5/d}$ (Eq.~\eqref{eq:pca-init}).
Because this rescales the baselines' intended initialization magnitudes, we
audited the two baselines for which magnitude is most central at $N=64$ with
three seeds (Table~\ref{tab:baseline-fidelity}). The reference PEFT
implementation of PiSSA matches our implementation to within 0.012 on all three
tasks. Restoring LoRAM's intended $\log(1+\sigma_{\max})$ magnitude changes
its score by $+0.019$, $-0.017$, and $-0.035$, and it remains below vanilla LoRA
on every task. The rescaling therefore does not change the baseline
conclusions in Table~\ref{tab:bert-primary-results}, and the as-released runs
below leave PivotLoRA with the highest mean at every $N$.

\begin{table}[H]
\centering
\caption{Baseline-fidelity audit at $N=64$, mean over three seeds. Each block
is a separate matched run; compare within blocks. CoLA uses Matthews
correlation.}
\label{tab:baseline-fidelity}
\small
\begin{tabular}{@{}lccc@{}}
\toprule
Method & AG~News & SST-2 & CoLA \\
\midrule
LoRA & .634 & .633 & .183 \\
PiSSA (our implementation) & .477 & .552 & .099 \\
PiSSA (reference PEFT) & .472 & .548 & .087 \\
\midrule
LoRA & .636 & .639 & .201 \\
LoRAM (our implementation) & .476 & .597 & .141 \\
LoRAM (intended magnitude) & .495 & .580 & .106 \\
\bottomrule
\end{tabular}
\end{table}

\paragraph{Baselines as released.}
\label{app:baselines-released}
To check that the shared rescaling does not favor PivotLoRA, we also ran
LoRAM, LoRA-GA, LoRA-One, and SRLoRA as released, following each method's
official initialization and scaling, and SRLoRA's published schedule, without
the shared $\sqrt{5/d}$ factor, together with vanilla LoRA under the standard PEFT
initialization ($A\in\mathbb{R}^{r\times d}$, $\|A\|_F\approx\sqrt{r/3}$). All
runs use the matched BERT protocol below with three seeds
(Table~\ref{tab:baselines-released}). LoRA-GA as released is stronger than our
version, but PivotLoRA keeps the highest mean at every $N$ against every
baseline in this set.

\begin{table}[H]
\centering
\caption{BERT-base baselines run as released, mean over six tasks and three
seeds. The last column counts the 30 task--$N$ settings in which PivotLoRA
has the higher mean.}
\label{tab:baselines-released}
\small
\begin{tabular}{@{}lcccccc@{}}
\toprule
Method & 16 & 32 & 64 & 128 & 256 & PivotLoRA ahead \\
\midrule
LoRA (as in Table~\ref{tab:bert-primary-results}) & .402 & .428 & .488 & .566 & .618 & 29/30 \\
LoRA (standard PEFT initialization) & .385 & .391 & .438 & .501 & .599 & 28/30 \\
LoRAM (as released) & .390 & .400 & .433 & .487 & .573 & 29/30 \\
LoRA-GA (as released) & .427 & .452 & .523 & .583 & .636 & 24/30 \\
LoRA-One (as released) & .409 & .417 & .462 & .575 & .621 & 28/30 \\
SRLoRA (as released) & .389 & .394 & .427 & .478 & .555 & 30/30 \\
\midrule
PivotLoRA & .451 & .500 & .575 & .606 & .641 & -- \\
\bottomrule
\end{tabular}
\end{table}

\paragraph{Matched BERT protocol.}
The main BERT comparison uses rank $r=8$, AdamW, 15 epochs, batch size 16,
query and value adapters, and a shared base learning-rate anchor of
$\eta_A=2\times10^{-4}$. Unless a baseline's standard parameterization
requires otherwise, $\alpha_{\mathrm{LoRA}}=16$. The learning-rate sweeps in
Appendix~\ref{app:lr-sweep} test sensitivity to this anchor.
Every method is evaluated after each epoch, and we report each run's best
evaluation score over the 15 epochs, using the same rule for all methods;
CoLA's Matthews correlation is floored at 0.

\subsection{Classification Learning-Rate Allocation}
\label{app:classification-lr}

Here, $\eta_A$ denotes the base learning-rate anchor. Methods with shared
factor-wise rates apply this rate directly to both LoRA factors, whereas
methods with differential rates derive their $B$-factor rates from it. At
initialization, TT-PCA and PivotLoRA assign the rank-$i$ coordinate of $B$ a
learning rate proportional to the corresponding PCA eigenvalue:
\begin{equation}
\label{eq:eigenlr}
  \eta_{B,i}
  =
  \frac{\lambda_i}{\bar\lambda}\,
  \lambda_{\mathrm{ratio}}\eta_A,
  \qquad
  \bar\lambda
  =
  \frac{1}{r}\sum_{j=1}^{r}\lambda_j,
  \qquad
  \lambda_{\mathrm{ratio}}=16.
\end{equation}
Normalizing by $\bar\lambda$ makes the average $B$-to-$A$ learning-rate ratio
equal to 16. This allocation is an optimizer heuristic and is not part of the activation-subspace definition. The controls in
Appendix~\ref{app:direction-lr-controls} separate the selected directions
from the eigenvalue-to-coordinate rate assignment.

TT-PCA uses these rates throughout training. PivotLoRA uses them except for the 10 optimizer steps after the pivot, during which a common $B$ rate rises from $\eta_A$ to $16\eta_A$ as a precaution, because the pivot resets the Adam state of $A$ while the state of $B$ still refers to the previous coordinates. Appendix~\ref{app:post-pivot-lr} shows that this precaution is not needed. Post-pivot learning-rate policies differ by at most 0.011 in average score, and the pivot with no learning-rate change performs as well. The reasoning experiments instead use a single learning rate
of $2\times10^{-4}$ for both factors, with no differential or eigenvalue-scaled rates and no post-pivot change.

\subsection{TT-LDA Ablation}
\label{app:tt-lda}

TT-LDA uses the same prediction-position representations as TT-PCA but
constructs supervised linear-discriminant directions. Because a $C$-class
problem supplies at most $C-1$ discriminant directions, residual PCA
directions fill the unused coordinates when $C-1<r$. The combined basis is
orthogonalized to rank $r=8$. TT-LDA uses the uniform rate $\eta_B=16\eta_A$, with no eigenvalue scaling across ranks. We use TT-LDA as an ablation of
the activation statistic, not as the default method. Comparative results
appear in Appendix~\ref{app:lr-structure}.

\subsection{Classification Pivot Configuration}
\label{app:classification-pivot}

The primary classification configuration uses pivot epoch $K=3$. At the
pivot, reduced QR is applied to $A^\top$ in float32, and $B$ is re-encoded according to
Eq.~\eqref{eq:reproj}. Adam first- and second-moment states for $A$ are reset.
The state for $B$ is retained, and its learning rate is handled as described in Appendix~\ref{app:classification-lr}. The reset policy is an
implementation choice and is separate from the algebraic output-preservation
guarantee.

\begin{algorithm}[H]
\centering
\small
\caption{Classification PivotLoRA}
\label{alg:pivotlora}
\begin{algorithmic}[1]
\Require Frozen model $\mathcal M$, labeled data $\mathcal D$, rank $r$,
pivot epoch $K$
\State Collect prediction-position activations $X_\ell$ at each adapted layer
\For{each adapted layer $\ell$}
  \State Center $X_\ell$ and compute its top-$r$ PCA basis
  \State Initialize $\operatorname{row}(A_\ell)$ from the PCA basis
  \State Set $B_\ell\gets0$
  \State Assign the $B_\ell$ coordinate rates using
    Eq.~\eqref{eq:eigenlr}
\EndFor
\State Train through epoch $K$
\For{each adapted layer $\ell$}
  \State $A_{\mathrm{old}}\gets A_\ell$ and
    $B_{\mathrm{old}}\gets B_\ell$
  \State $Q,R\gets\operatorname{QR}(A_{\mathrm{old}}^\top)$
  \State $A_{\mathrm{new}}\gets Q^\top$
  \State
    $B_{\mathrm{new}}\gets
     B_{\mathrm{old}}A_{\mathrm{old}}A_{\mathrm{new}}^\top$
  \State Replace $(A_\ell,B_\ell)$ by
    $(A_{\mathrm{new}},B_{\mathrm{new}})$
  \State Reset the Adam state for $A_\ell$
\EndFor
\State Ramp the $B$ rate from $\eta_A$ to $16\eta_A$ over 10 steps, then resume Eq.~\eqref{eq:eigenlr}
\State Train for the remaining epochs
\end{algorithmic}
\end{algorithm}

\paragraph{GPT-2 summarization configuration.}
The loss is applied only to the supervised summary, and the tied language-modeling head remains
frozen. We cross two activation sources with two learning-rate recipes. The sources are the mean over all positions of the padded sequence and the mean over response positions (those predicting supervised summary tokens). The recipes are the classification recipe of Eq.~\eqref{eq:eigenlr} with $\lambda_{\mathrm{ratio}}=16$ and a single rate of $2\times10^{-4}$ for both factors, as in the reasoning experiments (Appendix~\ref{app:generation}).

\FloatBarrier

\section{Generated-Answer Reasoning Details}
\label{app:reasoning-details}

We evaluate response-position adaptation on GSM8K and StrategyQA. LoRA
and PivotLoRA use matched data, training, and evaluation protocols, and PivotLoRA
additionally uses activation-based initialization and one output-preserving
pivot. Exact commands and lower-level implementation details are provided with
the released code.

StrategyQA was selected with a screening step fixed in advance that ran only the base model and LoRA. A candidate dataset was kept if
LoRA was clearly above chance, followed the trained answer format in at least
90\% of generations, and rarely hit the generation limit. No PivotLoRA run was
made on a candidate before it passed. We also considered MATH-500 but stopped at this stage. Across three pipeline checks, each method solved between 1 and
10 of 50 problems in our 128-example fine-tuning setting, and fewer than 80\%
of trained-model generations followed the boxed-answer format, so a comparison
would have been dominated by noise and format failures.

\subsection{Data and Response-Position Activations}
\label{app:reasoning-data}

GSM8K contains 7,473 training examples and 1,319 test examples, and StrategyQA
(the \texttt{ChilleD/StrategyQA} split) contains 1,603 training examples and
687 test examples. For StrategyQA, the one training question that also appears in the
test split is removed from the training pool. For each dataset, we
sample $N=128$ training examples without replacement using seeds
$\{42,43,44\}$. LoRA and PivotLoRA use the same subset at each seed.

The GSM8K prompt is
\begin{verbatim}
Question: {question}
Answer:
\end{verbatim}
with a single space followed by the dataset's original answer field as the
completion. StrategyQA uses the GSM8K prompt
with completion
\begin{verbatim}
 {facts}
#### {yes|no}
\end{verbatim}
where \texttt{facts} is the dataset's supporting-facts field.

Each prompt ends at \texttt{Answer:} and each completion begins with a space,
so the prompt used at inference is an exact token-level prefix of every
training sequence, which the implementation asserts for every training example.
Generation therefore starts from the same token boundary seen in training.

Prompt tokens are excluded from the language-modeling loss, whereas completion
tokens and the appended EOS token are supervised. Sequences are jointly
tokenized and truncated to a maximum length of 512 tokens. Because hidden
state $t$ predicts token $t+1$, the response-predictor positions for example
$n$ are

\[
  \mathcal{P}_n
  =
  \{t : M_{n,t+1}=1\},
\]
where $M$ denotes the completion mask. At adapted layer $\ell$, the activation
used for PCA is

\[
  x_{\ell,n}
  =
  \frac{1}{|\mathcal{P}_n|}
  \sum_{t\in\mathcal{P}_n} h_{\ell,n,t}.
\]
The resulting example-level activation matrix is centered before PCA.
Averaging within each example prevents response length from determining its
weight in the covariance estimate.

\subsection{Training and Pivot Configuration}
\label{app:reasoning-config}

\begin{table}[H]
\centering
\caption{Configuration for the reasoning experiments, shared by both
datasets.}
\label{tab:reasoning-config}
\small
\setlength{\tabcolsep}{4pt}
\begin{tabular}{@{}ll@{}}
\toprule
Field & Value \\
\midrule
Checkpoint & \texttt{meta-llama/Llama-3.1-8B} \\
Target modules & \texttt{q\_proj}, \texttt{v\_proj} \\
Adapted modules & 64 \\
Rank & $r=8$ \\
LoRA scaling & $\alpha_{\mathrm{LoRA}}=16$ \\
Optimizer & AdamW \\
Learning rate & $2\times10^{-4}$ \\
Weight decay & 0.01 \\
Schedule & Linear decay, no warm-up \\
Epochs & 15 \\
Per-device batch size & 4 \\
Gradient accumulation & 2 \\
Effective batch size & 8 \\
Optimizer steps & 240 \\
Maximum sequence length & 512 \\
Base-model precision & bf16 \\
Adapter precision & float32 \\
Pivot epoch & 3 \\
Pivot step & 48 \\
\bottomrule
\end{tabular}
\end{table}

The PCA basis is orthogonalized and scaled by $\sqrt{5/d}$, with $d=4096$,
before assignment to the LoRA input factor $A$, with no random rotation.
PCA is computed in float32. PivotLoRA performs one
pivot at epoch 3, corresponding to global step 48, using float64 QR, and the transformed factors are stored in float32. All 64 query and value
adapters are transformed.

Immediately after the pivot, the AdamW states for both $A$ and $B$ are reset,
while the learning-rate schedule continues from the current global step. Resetting both states differs from the classification policy (Appendix~\ref{app:classification-pivot}).

The $\sqrt{5/d}$ basis scale gives $A$ about $17\times$ smaller Frobenius norm
than vanilla Kaiming-uniform initialization, so it is not norm-matched. These experiments therefore evaluate the complete initialization-and-pivot procedure, not direction in isolation.

\subsection{Generation and Evaluation}
\label{app:reasoning-eval}

Evaluation uses greedy decoding with batch size 1 and at most 512 new tokens,
terminating earlier at EOS. The model uses EOS for padding and
\texttt{sdpa} attention.

The evaluator first extracts a numeric value following the final
\texttt{\#\#\#\#} delimiter. If none is present, it uses the final numeric
value in the generated response. Predictions and references are normalized as
numeric strings and compared by exact equality. Signed integers,
comma-grouped numbers, and decimals are supported; fractions and percentages
are not converted. Outputs without a valid numeric value are counted as
incorrect. For StrategyQA, the evaluator instead extracts the first standalone
\texttt{yes} or \texttt{no} after the final \texttt{\#\#\#\#}, falling back
to the last standalone \texttt{yes} or \texttt{no} in the response, and
compares it with the label.

We additionally report format compliance, the fraction of generations that
follow the trained answer format, which ends in \texttt{\#\#\#\# <number>} for
GSM8K and in \texttt{\#\#\#\# yes} or \texttt{\#\#\#\# no} for
StrategyQA. As a reference, the frozen base model
without adapters is evaluated once with the same prompt, decoding, and scorer.

\subsection{Per-Seed Results}
\label{app:reasoning-results}

\begin{table}[H]
\centering
\caption{GSM8K results. Accuracy is numeric exact match on the 1,319-example
test set. Format is the percentage of generations following the trained answer
format. The base model has no adapter and is evaluated once.}
\label{tab:gsm8k-per-seed}
\small
\begin{tabular}{@{}lrrrrr@{}}
\toprule
Method & Seed & Correct & Accuracy (\%) & Parse rate (\%) & Format (\%) \\
\midrule
Base model & -- & 338 & 25.63 & 100.00 & 0.0 \\
\midrule
LoRA      & 42 & 376 & 28.51 & 100.00 & 98.0 \\
LoRA      & 43 & 381 & 28.89 & 100.00 & 96.0 \\
LoRA      & 44 & 378 & 28.66 & 100.00 & 95.8 \\
PivotLoRA & 42 & 421 & 31.92 & 100.00 & 98.2 \\
PivotLoRA & 43 & 476 & 36.09 & 100.00 & 97.6 \\
PivotLoRA & 44 & 418 & 31.69 & 100.00 & 97.4 \\
\bottomrule
\end{tabular}
\end{table}

PivotLoRA improves on LoRA at every seed, by $+3.41$, $+7.20$, and $+3.03$ points (mean $+4.55\pm2.31$, sample SD over seeds), and even the smallest gain exceeds LoRA's own seed-to-seed spread ($\pm0.19$). Pooled over the three seeds, PivotLoRA alone answers 655 test items correctly against 475 for LoRA alone (exact McNemar $p=9.5\times10^{-8}$), and the test is significant at each seed ($p\le0.036$).

\begin{table}[H]
\centering
\caption{StrategyQA results. Accuracy is yes/no exact match on the
687-example test split. Columns as in Table~\ref{tab:gsm8k-per-seed}.}
\label{tab:strategyqa-per-seed}
\small
\begin{tabular}{@{}lrrrrr@{}}
\toprule
Method & Seed & Correct & Accuracy (\%) & Parse rate (\%) & Format (\%) \\
\midrule
Base model & -- & 374 & 54.44 & 85.15 & 0.0 \\
\midrule
LoRA      & 42 & 439 & 63.90 & 98.11 & 95.2 \\
LoRA      & 43 & 457 & 66.52 & 99.85 & 99.4 \\
LoRA      & 44 & 426 & 62.01 & 99.42 & 96.7 \\
PivotLoRA & 42 & 476 & 69.29 & 99.85 & 99.6 \\
PivotLoRA & 43 & 457 & 66.52 & 99.71 & 99.4 \\
PivotLoRA & 44 & 463 & 67.39 & 99.85 & 98.0 \\
\bottomrule
\end{tabular}
\end{table}

PivotLoRA improves on LoRA by $+5.39$ points at two seeds and ties at the third (mean $+3.59\pm3.11$). The tie does not reflect identical outputs, since each method alone answers 98 questions correctly. Pooled over the three seeds, PivotLoRA alone answers 350 test items correctly against 276 for LoRA alone (exact McNemar $p=0.004$).

\section{Extended Experimental Results}
\label{app:results}

\subsection{Complete BERT-Base Results}
\label{app:fullresults}

Tables~\ref{tab:app-sst2}--\ref{tab:app-boolq} report mean and sample standard
deviation over three seeds. The six primary tasks contain 30 task--$N$
settings. BoolQ is an auxiliary classification task used in the alignment
analysis. All methods follow the shared benchmark protocol together with the
method-specific configurations documented in
Appendix~\ref{app:implementation}.

\begin{table}[H]
\centering
\caption{SST-2 accuracy, reported as mean $\pm$ standard deviation over three
seeds. Columns denote the total number of labeled examples $N$.}
\label{tab:app-sst2}
\small
\setlength{\tabcolsep}{3pt}
\begin{tabular}{@{}lccccc@{}}
\toprule
Method & 16 & 32 & 64 & 128 & 256 \\
\midrule
LoRA
& .524$\pm$.011 & .546$\pm$.037 & .652$\pm$.014
& .772$\pm$.006 & .844$\pm$.012 \\
DoRA
& .528$\pm$.014 & .544$\pm$.036 & .647$\pm$.018
& .761$\pm$.010 & .841$\pm$.008 \\
AdaLoRA
& .502$\pm$.019 & .500$\pm$.008 & .548$\pm$.016
& .580$\pm$.026 & .633$\pm$.019 \\
PiSSA
& .503$\pm$.023 & .500$\pm$.008 & .548$\pm$.027
& .665$\pm$.058 & .823$\pm$.002 \\
LoRA-GA
& .503$\pm$.018 & .500$\pm$.007 & .565$\pm$.033
& .664$\pm$.077 & .849$\pm$.004 \\
LoRA-One
& .497$\pm$.015 & .501$\pm$.007 & .565$\pm$.032
& .711$\pm$.059 & .854$\pm$.010 \\
LoRA+
& .551$\pm$.033 & .569$\pm$.042 & .538$\pm$.038
& .509$\pm$.000 & .509$\pm$.000 \\
EVA
& .503$\pm$.017 & .500$\pm$.007 & .561$\pm$.031
& .683$\pm$.032 & .855$\pm$.010 \\
LoRAM
& .498$\pm$.015 & .500$\pm$.007 & .568$\pm$.030
& .702$\pm$.060 & .849$\pm$.005 \\
SRLoRA
& .524$\pm$.011 & .536$\pm$.036 & .632$\pm$.014
& .750$\pm$.014 & .843$\pm$.011 \\
\midrule
TT-PCA
& .520$\pm$.024 & .530$\pm$.011 & .755$\pm$.007
& .817$\pm$.003 & .862$\pm$.007 \\
TT-LDA
& .523$\pm$.020 & .542$\pm$.014 & .783$\pm$.007
& .828$\pm$.014 & .859$\pm$.011 \\
PivotLoRA
& .543$\pm$.038 & .633$\pm$.018 & .795$\pm$.016
& .823$\pm$.017 & .850$\pm$.010 \\
\bottomrule
\end{tabular}
\end{table}
\FloatBarrier

\begin{table}[H]
\centering
\caption{CoLA Matthews correlation, reported as mean $\pm$ standard deviation
over three seeds.}
\label{tab:app-cola}
\small
\setlength{\tabcolsep}{3pt}
\begin{tabular}{@{}lccccc@{}}
\toprule
Method & 16 & 32 & 64 & 128 & 256 \\
\midrule
LoRA
& .081$\pm$.056 & .131$\pm$.047 & .163$\pm$.046
& .264$\pm$.072 & .407$\pm$.010 \\
DoRA
& .085$\pm$.060 & .128$\pm$.041 & .171$\pm$.042
& .257$\pm$.070 & .393$\pm$.007 \\
AdaLoRA
& .096$\pm$.055 & .081$\pm$.040 & .100$\pm$.029
& .152$\pm$.033 & .183$\pm$.023 \\
PiSSA
& .090$\pm$.064 & .111$\pm$.040 & .111$\pm$.083
& .206$\pm$.059 & .338$\pm$.022 \\
LoRA-GA
& .086$\pm$.061 & .097$\pm$.045 & .103$\pm$.078
& .214$\pm$.059 & .382$\pm$.003 \\
LoRA-One
& .087$\pm$.060 & .089$\pm$.039 & .111$\pm$.062
& .240$\pm$.067 & .387$\pm$.007 \\
LoRA+
& .071$\pm$.023 & .105$\pm$.041 & .038$\pm$.012
& .000$\pm$.000 & .000$\pm$.000 \\
EVA
& .089$\pm$.061 & .095$\pm$.036 & .101$\pm$.070
& .243$\pm$.058 & .387$\pm$.017 \\
LoRAM
& .089$\pm$.061 & .085$\pm$.036 & .112$\pm$.070
& .239$\pm$.058 & .390$\pm$.017 \\
SRLoRA
& .081$\pm$.058 & .129$\pm$.030 & .161$\pm$.051
& .240$\pm$.065 & .405$\pm$.008 \\
\midrule
TT-PCA
& .103$\pm$.044 & .137$\pm$.046 & .274$\pm$.094
& .381$\pm$.049 & .397$\pm$.019 \\
TT-LDA
& .106$\pm$.036 & .150$\pm$.041 & .281$\pm$.100
& .377$\pm$.071 & .412$\pm$.016 \\
PivotLoRA
& .124$\pm$.015 & .171$\pm$.049 & .288$\pm$.098
& .365$\pm$.071 & .411$\pm$.025 \\
\bottomrule
\end{tabular}
\end{table}
\FloatBarrier

\begin{table}[H]
\centering
\caption{MNLI matched-validation accuracy, reported as mean $\pm$ standard
deviation over three seeds.}
\label{tab:app-mnli}
\small
\setlength{\tabcolsep}{3pt}
\begin{tabular}{@{}lccccc@{}}
\toprule
Method & 16 & 32 & 64 & 128 & 256 \\
\midrule
LoRA
& .350$\pm$.008 & .356$\pm$.006 & .366$\pm$.003
& .356$\pm$.008 & .392$\pm$.011 \\
DoRA
& .349$\pm$.008 & .355$\pm$.006 & .366$\pm$.002
& .359$\pm$.008 & .392$\pm$.012 \\
AdaLoRA
& .350$\pm$.004 & .354$\pm$.004 & .347$\pm$.013
& .352$\pm$.009 & .356$\pm$.013 \\
PiSSA
& .352$\pm$.003 & .354$\pm$.002 & .349$\pm$.014
& .354$\pm$.009 & .383$\pm$.023 \\
LoRA-GA
& .350$\pm$.003 & .356$\pm$.003 & .348$\pm$.013
& .351$\pm$.009 & .397$\pm$.014 \\
LoRA-One
& .350$\pm$.004 & .355$\pm$.004 & .348$\pm$.013
& .357$\pm$.010 & .401$\pm$.012 \\
LoRA+
& .353$\pm$.012 & .355$\pm$.001 & .352$\pm$.019
& .355$\pm$.000 & .355$\pm$.000 \\
EVA
& .350$\pm$.004 & .356$\pm$.003 & .347$\pm$.014
& .352$\pm$.008 & .397$\pm$.015 \\
LoRAM
& .350$\pm$.004 & .356$\pm$.003 & .349$\pm$.014
& .356$\pm$.008 & .404$\pm$.015 \\
SRLoRA
& .349$\pm$.007 & .356$\pm$.005 & .367$\pm$.003
& .357$\pm$.007 & .390$\pm$.009 \\
\midrule
TT-PCA
& .349$\pm$.005 & .359$\pm$.007 & .367$\pm$.003
& .378$\pm$.011 & .424$\pm$.025 \\
TT-LDA
& .352$\pm$.007 & .357$\pm$.005 & .366$\pm$.003
& .379$\pm$.007 & .425$\pm$.024 \\
PivotLoRA
& .356$\pm$.010 & .362$\pm$.008 & .367$\pm$.005
& .389$\pm$.009 & .454$\pm$.041 \\
\bottomrule
\end{tabular}
\end{table}
\FloatBarrier

\begin{table}[H]
\centering
\caption{QNLI accuracy, reported as mean $\pm$ standard deviation over three
seeds.}
\label{tab:app-qnli}
\small
\setlength{\tabcolsep}{3pt}
\begin{tabular}{@{}lccccc@{}}
\toprule
Method & 16 & 32 & 64 & 128 & 256 \\
\midrule
LoRA
& .519$\pm$.037 & .521$\pm$.040 & .555$\pm$.042
& .626$\pm$.061 & .662$\pm$.040 \\
DoRA
& .517$\pm$.039 & .519$\pm$.040 & .556$\pm$.042
& .622$\pm$.055 & .663$\pm$.035 \\
AdaLoRA
& .513$\pm$.029 & .514$\pm$.029 & .527$\pm$.023
& .537$\pm$.029 & .546$\pm$.024 \\
PiSSA
& .510$\pm$.027 & .512$\pm$.041 & .528$\pm$.021
& .574$\pm$.035 & .616$\pm$.038 \\
LoRA-GA
& .512$\pm$.027 & .514$\pm$.035 & .528$\pm$.024
& .567$\pm$.045 & .641$\pm$.043 \\
LoRA-One
& .511$\pm$.028 & .516$\pm$.031 & .529$\pm$.023
& .574$\pm$.049 & .644$\pm$.045 \\
LoRA+
& .539$\pm$.033 & .576$\pm$.052 & .554$\pm$.040
& .515$\pm$.012 & .505$\pm$.000 \\
EVA
& .511$\pm$.028 & .516$\pm$.031 & .532$\pm$.024
& .575$\pm$.052 & .645$\pm$.050 \\
LoRAM
& .512$\pm$.028 & .516$\pm$.031 & .530$\pm$.024
& .567$\pm$.052 & .644$\pm$.050 \\
SRLoRA
& .512$\pm$.038 & .519$\pm$.039 & .551$\pm$.040
& .618$\pm$.059 & .662$\pm$.034 \\
\midrule
TT-PCA
& .517$\pm$.034 & .525$\pm$.031 & .574$\pm$.059
& .640$\pm$.057 & .677$\pm$.024 \\
TT-LDA
& .517$\pm$.035 & .528$\pm$.033 & .572$\pm$.048
& .638$\pm$.055 & .681$\pm$.029 \\
PivotLoRA
& .533$\pm$.026 & .547$\pm$.036 & .594$\pm$.049
& .645$\pm$.050 & .691$\pm$.013 \\
\bottomrule
\end{tabular}
\end{table}
\FloatBarrier

\begin{table}[H]
\centering
\caption{RTE accuracy, reported as mean $\pm$ standard deviation over three
seeds.}
\label{tab:app-rte}
\small
\setlength{\tabcolsep}{3pt}
\begin{tabular}{@{}lccccc@{}}
\toprule
Method & 16 & 32 & 64 & 128 & 256 \\
\midrule
LoRA
& .508$\pm$.040 & .549$\pm$.036 & .551$\pm$.022
& .569$\pm$.015 & .549$\pm$.003 \\
DoRA
& .511$\pm$.040 & .546$\pm$.024 & .555$\pm$.015
& .568$\pm$.018 & .540$\pm$.007 \\
AdaLoRA
& .510$\pm$.035 & .526$\pm$.011 & .551$\pm$.022
& .540$\pm$.012 & .524$\pm$.014 \\
PiSSA
& .505$\pm$.033 & .528$\pm$.032 & .539$\pm$.019
& .555$\pm$.025 & .528$\pm$.015 \\
LoRA-GA
& .516$\pm$.036 & .525$\pm$.018 & .552$\pm$.016
& .554$\pm$.015 & .527$\pm$.014 \\
LoRA-One
& .515$\pm$.037 & .528$\pm$.017 & .561$\pm$.021
& .561$\pm$.007 & .540$\pm$.013 \\
LoRA+
& .545$\pm$.028 & .552$\pm$.011 & .527$\pm$.000
& .527$\pm$.000 & .527$\pm$.000 \\
EVA
& .513$\pm$.037 & .534$\pm$.024 & .555$\pm$.017
& .563$\pm$.013 & .539$\pm$.024 \\
LoRAM
& .513$\pm$.037 & .530$\pm$.024 & .563$\pm$.017
& .566$\pm$.013 & .551$\pm$.024 \\
SRLoRA
& .513$\pm$.046 & .552$\pm$.031 & .539$\pm$.021
& .567$\pm$.018 & .549$\pm$.009 \\
\midrule
TT-PCA
& .522$\pm$.051 & .563$\pm$.033 & .555$\pm$.025
& .561$\pm$.022 & .552$\pm$.005 \\
TT-LDA
& .516$\pm$.055 & .567$\pm$.023 & .558$\pm$.020
& .561$\pm$.028 & .555$\pm$.007 \\
PivotLoRA
& .551$\pm$.019 & .584$\pm$.017 & .569$\pm$.025
& .549$\pm$.010 & .563$\pm$.012 \\
\bottomrule
\end{tabular}
\end{table}
\FloatBarrier

\begin{table}[H]
\centering
\caption{AG~News accuracy, reported as mean $\pm$ standard deviation over
three seeds.}
\label{tab:app-agnews}
\small
\setlength{\tabcolsep}{3pt}
\begin{tabular}{@{}lccccc@{}}
\toprule
Method & 16 & 32 & 64 & 128 & 256 \\
\midrule
LoRA
& .431$\pm$.016 & .466$\pm$.050 & .640$\pm$.033
& .807$\pm$.006 & .855$\pm$.006 \\
DoRA
& .432$\pm$.015 & .446$\pm$.041 & .619$\pm$.016
& .801$\pm$.010 & .857$\pm$.010 \\
AdaLoRA
& .304$\pm$.027 & .308$\pm$.016 & .368$\pm$.038
& .438$\pm$.031 & .504$\pm$.028 \\
PiSSA
& .338$\pm$.031 & .364$\pm$.007 & .479$\pm$.024
& .643$\pm$.036 & .795$\pm$.024 \\
LoRA-GA
& .330$\pm$.022 & .354$\pm$.012 & .466$\pm$.038
& .679$\pm$.042 & .812$\pm$.030 \\
LoRA-One
& .314$\pm$.025 & .338$\pm$.012 & .475$\pm$.033
& .696$\pm$.055 & .808$\pm$.032 \\
LoRA+
& .695$\pm$.011 & .730$\pm$.044 & .619$\pm$.136
& .646$\pm$.281 & .594$\pm$.123 \\
EVA
& .317$\pm$.024 & .345$\pm$.012 & .481$\pm$.040
& .702$\pm$.070 & .815$\pm$.029 \\
LoRAM
& .311$\pm$.024 & .333$\pm$.012 & .482$\pm$.040
& .672$\pm$.070 & .809$\pm$.029 \\
SRLoRA
& .426$\pm$.021 & .446$\pm$.045 & .584$\pm$.015
& .795$\pm$.013 & .851$\pm$.004 \\
\midrule
TT-PCA
& .478$\pm$.028 & .558$\pm$.012 & .753$\pm$.041
& .858$\pm$.005 & .874$\pm$.006 \\
TT-LDA
& .475$\pm$.026 & .587$\pm$.015 & .769$\pm$.027
& .861$\pm$.009 & .871$\pm$.008 \\
PivotLoRA
& .600$\pm$.049 & .705$\pm$.047 & .834$\pm$.018
& .866$\pm$.004 & .878$\pm$.001 \\
\bottomrule
\end{tabular}
\end{table}
\FloatBarrier

\begin{table}[H]
\centering
\caption{BoolQ accuracy, reported as mean $\pm$ standard deviation over three
seeds.}
\label{tab:app-boolq}
\small
\setlength{\tabcolsep}{3pt}
\begin{tabular}{@{}lccccc@{}}
\toprule
Method & 16 & 32 & 64 & 128 & 256 \\
\midrule
LoRA
& .602$\pm$.005 & .603$\pm$.010 & .551$\pm$.048
& .567$\pm$.039 & .601$\pm$.019 \\
DoRA
& .601$\pm$.005 & .606$\pm$.007 & .553$\pm$.043
& .573$\pm$.032 & .596$\pm$.022 \\
AdaLoRA
& .593$\pm$.018 & .603$\pm$.012 & .549$\pm$.030
& .542$\pm$.024 & .576$\pm$.017 \\
PiSSA
& .597$\pm$.015 & .603$\pm$.012 & .549$\pm$.029
& .563$\pm$.030 & .591$\pm$.035 \\
LoRA-GA
& .596$\pm$.015 & .602$\pm$.011 & .565$\pm$.006
& .562$\pm$.030 & .584$\pm$.017 \\
LoRA-One
& .593$\pm$.018 & .602$\pm$.011 & .538$\pm$.032
& .549$\pm$.010 & .580$\pm$.017 \\
LoRA+
& .591$\pm$.042 & .622$\pm$.000 & .621$\pm$.001
& .622$\pm$.000 & .622$\pm$.000 \\
EVA
& .594$\pm$.017 & .603$\pm$.012 & .537$\pm$.033
& .538$\pm$.014 & .579$\pm$.020 \\
LoRAM
& .593$\pm$.018 & .603$\pm$.012 & .545$\pm$.029
& .549$\pm$.009 & .576$\pm$.017 \\
SRLoRA
& .602$\pm$.005 & .603$\pm$.010 & .549$\pm$.050
& .566$\pm$.039 & .599$\pm$.021 \\
\midrule
TT-PCA
& .596$\pm$.016 & .601$\pm$.014 & .568$\pm$.014
& .574$\pm$.006 & .612$\pm$.010 \\
TT-LDA
& .596$\pm$.016 & .602$\pm$.013 & .569$\pm$.016
& .567$\pm$.009 & .612$\pm$.005 \\
PivotLoRA
& .596$\pm$.016 & .601$\pm$.014 & .567$\pm$.029
& .589$\pm$.011 & .611$\pm$.010 \\
\bottomrule
\end{tabular}
\end{table}
\FloatBarrier

\subsection{Full Aggregate Rankings}
\label{app:fullrankings}

Table~\ref{tab:fullrankings} aggregates the 30 primary BERT task--$N$
settings. PivotLoRA has average rank 1.83 across all 13 methods and has the strictly highest mean in 19 of 30 settings, counting TT-PCA and TT-LDA, and ties for highest on MNLI at \(N=64\). It
exceeds vanilla LoRA in 29 of the 30 settings. 

\begin{table}[H]
\centering
\caption{Aggregate statistics across the 30 primary BERT-base few-shot
settings. ``Wins vs.\ PivotLoRA'' counts settings in which the listed method has
a strictly higher mean than PivotLoRA.}
\label{tab:fullrankings}
\small
\begin{tabular}{@{}lcc@{}}
\toprule
Method & Average rank ($\downarrow$) & Wins vs.\ PivotLoRA \\
\midrule
PivotLoRA (Ours) & \textbf{1.83} & N/A \\
TT-LDA           & 3.13 & 5/30 \\
TT-PCA           & 3.77 & 3/30 \\
LoRA             & 5.23 & 1/30 \\
DoRA             & 5.83 & 1/30 \\
SRLoRA           & 6.93 & 1/30 \\
LoRA+            & 8.43 & 5/30 \\
LoRAM            & 8.43 & 1/30 \\
LoRA-One         & 8.47 & 2/30 \\
EVA              & 8.47 & 2/30 \\
LoRA-GA          & 9.30 & 1/30 \\
PiSSA            & 9.87 & 1/30 \\
AdaLoRA          & 11.30 & 0/30 \\
\bottomrule
\end{tabular}
\end{table}
\FloatBarrier

\subsection{Decoder-Scale Classification Results}
\label{app:decoder}

The decoder-scale experiments test whether the activation-subspace pattern
extends beyond BERT-base. Tables~\ref{tab:app-7b-llama2} and
\ref{tab:app-7b-llama31} report SST-2 and AG~News, while
Table~\ref{tab:app-7b-boolq} reports the BoolQ extension. PivotLoRA ranks first
at $N=64$ on SST-2 and AG~News for both backbones. At $N=256$, the
differences are smaller and PivotLoRA is not uniformly best.

EVA is included as an activation-based baseline under the same per-layer rank budget. It treats every token activation as a separate sample, whereas PivotLoRA uses one sequence-mean vector per example. 

\begin{table}[H]
\centering
\caption{Llama-2-7B few-shot classification, reported as mean $\pm$ standard
deviation over three seeds. Bold marks the
best mean among methods evaluated in the corresponding setting.}
\label{tab:app-7b-llama2}
\small
\setlength{\tabcolsep}{4pt}
\begin{tabular}{@{}lcccc@{}}
\toprule
Method & SST-2 64 & SST-2 256 & AG~News 64 & AG~News 256 \\
\midrule
LoRA
& .708$\pm$.051 & .924$\pm$.007
& .748$\pm$.027 & \textbf{.879}$\pm$.004 \\
EVA
& .713$\pm$.015 & .928$\pm$.016
& .745$\pm$.027 & \textbf{.879}$\pm$.004 \\
PiSSA
& .635$\pm$.058 & .922$\pm$.010
& .700$\pm$.014 & .865$\pm$.010 \\
LoRAM
& .763$\pm$.025 & \textbf{.929}$\pm$.014
& .742$\pm$.027 & .876$\pm$.003 \\
PivotLoRA
& \textbf{.769}$\pm$.030 & .912$\pm$.020
& \textbf{.784}$\pm$.029 & .861$\pm$.020 \\
\bottomrule
\end{tabular}
\end{table}
\FloatBarrier

\begin{table}[H]
\centering
\caption{Llama-3.1-8B few-shot classification, reported as mean $\pm$
sample standard deviation over three seeds. EVA entries at $N=64$ report
population standard deviations because per-seed values were not retained. Bold
marks the best mean among methods evaluated in the corresponding setting.}
\label{tab:app-7b-llama31}
\small
\setlength{\tabcolsep}{4pt}
\begin{tabular}{@{}lcccc@{}}
\toprule
Method & SST-2 64 & SST-2 256 & AG~News 64 & AG~News 256 \\
\midrule
LoRA
& .704$\pm$.037 & .883$\pm$.030
& .759$\pm$.023 & \textbf{.880}$\pm$.009 \\
EVA
& .643$\pm$.050 & \textbf{.909}$\pm$.017
& .743$\pm$.034 & .876$\pm$.009 \\
PiSSA
& .622$\pm$.043 & .876$\pm$.030
& .715$\pm$.027 & .864$\pm$.010 \\
LoRAM
& .662$\pm$.050 & .901$\pm$.012
& .748$\pm$.047 & .872$\pm$.010 \\
PivotLoRA
& \textbf{.709}$\pm$.085 & .899$\pm$.012
& \textbf{.799}$\pm$.001 & .870$\pm$.030 \\
\bottomrule
\end{tabular}
\end{table}
\FloatBarrier

\begin{table}[H]
\centering
\caption{Llama-3.1-8B classification on BoolQ, reported as mean $\pm$
population standard deviation over three seeds (per-seed values were not
retained).}
\label{tab:app-7b-boolq}
\small
\begin{tabular}{@{}lcc@{}}
\toprule
Method & $N=64$ & $N=256$ \\
\midrule
LoRA
& .554$\pm$.025 & .576$\pm$.020 \\
PiSSA
& .527$\pm$.023 & .536$\pm$.021 \\
LoRAM
& .534$\pm$.032 & .573$\pm$.014 \\
PivotLoRA
& \textbf{.602}$\pm$.009 & \textbf{.618}$\pm$.004 \\
\bottomrule
\end{tabular}
\end{table}
\FloatBarrier

\subsection{Summarization Control}
\label{app:generation}

GPT-2 summarization tests whether a large $d/N$ alone produces large gains
(Table~\ref{tab:app-generation}; XSum and CNN/DailyMail, ROUGE-L on 500
validation examples, three seeds, one matched sweep). Under the single learning
rate used for reasoning, PivotLoRA exceeds LoRA by only $+0.2$ points with the
sequence mean and $+0.3$ with response positions, winning 8 and 9 of the 12
seed-level comparisons, respectively. These gains are much smaller than the
classification and reasoning gains at comparable $d/N$. Under the
classification recipe, PivotLoRA trails LoRA by 1.5 to 5.1 points with either
source, so the learning-rate recipe matters more than the activation source. Response positions score above the sequence mean in all eight
comparisons, by $+0.1$ to $+0.2$ points under the single rate and by $+0.2$ to
$+1.2$ under the classification recipe.

\begin{table}[H]
\centering
\caption{GPT-2 few-shot summarization, ROUGE-L means over three seeds.
``classification'' uses eigenvalue-scaled rates with
$\lambda_{\mathrm{ratio}}=16$; ``single rate'' uses $2\times10^{-4}$ for both
factors, as in the reasoning experiments.}
\label{tab:app-generation}
\small
\begin{tabular}{@{}llcccc@{}}
\toprule
& & \multicolumn{2}{c}{XSum} & \multicolumn{2}{c}{CNN/DailyMail} \\
Method & Source / recipe & 64 & 256 & 64 & 256 \\
\midrule
LoRA & --- & .185 & .184 & \textbf{.212} & .227 \\
\midrule
PivotLoRA & sequence mean / classification & .135 & .164 & .193 & .208 \\
PivotLoRA & response / classification & .147 & .168 & .198 & .210 \\
PivotLoRA & sequence mean / single rate & .188 & .191 & .209 & .228 \\
PivotLoRA & response / single rate & \textbf{.191} & \textbf{.192} & .211 & \textbf{.229} \\
\bottomrule
\end{tabular}
\end{table}
\FloatBarrier

\subsection{Full-Data Reference}
\label{app:fulldata}

Table~\ref{tab:app-fulldata} compares PivotLoRA and LoRA when the full BERT
training split is used. The differences range from $-0.009$ to $+0.022$ in
the task's primary score. 

\begin{table}[H]
\centering
\caption{BERT-base on the full training split using seed 42. CoLA uses
Matthews correlation, and the remaining tasks use accuracy.}
\label{tab:app-fulldata}
\small
\begin{tabular}{@{}lcccccc@{}}
\toprule
Method & SST-2 & CoLA & MNLI & QNLI & RTE & AG~News \\
\midrule
LoRA
& .923 & .533 & .806 & .893 & .610 & .940 \\
PivotLoRA
& .914 & .555 & .817 & .889 & .610 & .944 \\
\midrule
$\Delta$
& $-.009$ & +.022 & +.011 & $-.004$ & .000 & +.004 \\
\bottomrule
\end{tabular}
\end{table}

\paragraph{PiSSA under full data.}
Because weight-spectral initializers such as PiSSA are often motivated by
full-data fine-tuning, we ran a separate matched single-seed audit on the full
training splits using the reference PEFT implementation of
PiSSA~\citep{meng2024pissa}. LoRA, PiSSA, and PivotLoRA share the data, seed,
optimizer, and training budget within this audit (Table~\ref{tab:app-pissa-full}).
Because it is a separate execution, its LoRA and PivotLoRA values differ
slightly from Table~\ref{tab:app-fulldata}. PiSSA is within $+0.018$ of LoRA on
five tasks and $0.021$ below it on CoLA, and PivotLoRA is at or above PiSSA on
all six tasks. With one seed, the three methods are comparable in the full-data regime. The audit does not measure convergence speed, where PiSSA reports its advantages.

\begin{table}[H]
\centering
\caption{Single-seed full-data audit with the reference PEFT implementation of
PiSSA (seed 42). CoLA uses Matthews correlation; the other tasks use accuracy.}
\label{tab:app-pissa-full}
\small
\begin{tabular}{@{}lcccccc@{}}
\toprule
Method & SST-2 & CoLA & MNLI & QNLI & RTE & AG~News \\
\midrule
LoRA      & .911 & .515 & .809 & .876 & .545 & .941 \\
PiSSA     & .915 & .494 & .812 & .879 & .563 & .941 \\
PivotLoRA & .916 & .565 & .815 & .902 & .606 & .943 \\
\bottomrule
\end{tabular}
\end{table}
\FloatBarrier

\subsection{Full Fine-Tuning Reference}
\label{app:fullft}

Table~\ref{tab:app-fullft} compares PivotLoRA with full fine-tuning at
$N=64$. Full fine-tuning is a strong reference but does not match the adapter
parameter budget. PivotLoRA exceeds the mean full fine-tuning score on all six tasks while training about \(370\times\) fewer parameters (approximately 0.30M). 

\begin{table}[H]
\centering
\caption{Full fine-tuning, LoRA, and PivotLoRA on BERT-base at $N=64$. Full
fine-tuning is reported as mean $\pm$ standard deviation over three seeds.
LoRA and PivotLoRA entries are the corresponding three-seed means from the
primary tables.}
\label{tab:app-fullft}
\small
\begin{tabular}{@{}lccc@{}}
\toprule
Task & Full FT (110M) & LoRA (0.30M) & PivotLoRA (0.30M) \\
\midrule
SST-2
& .700$\pm$.053 & .652 & \textbf{.795} \\
CoLA
& .286$\pm$.043 & .163 & \textbf{.288} \\
AG~News
& .791$\pm$.042 & .640 & \textbf{.834} \\
MNLI
& .345$\pm$.007 & .366 & \textbf{.367} \\
QNLI
& .555$\pm$.004 & .555 & \textbf{.594} \\
RTE
& .520$\pm$.046 & .551 & \textbf{.569} \\
\bottomrule
\end{tabular}
\end{table}
\FloatBarrier

\subsection{Alignment Estimation Procedure and Task-Level Gains}
\label{app:alignment}

For each BERT classification task, we run the frozen encoder on all $N$
labeled examples and extract final-layer \texttt{[CLS]} representations. The
few-shot set is sampled approximately class-balanced. Let \(n_c\) be the
number of examples in class \(c\), let \(\mu_c\) be its mean representation,
and let \(\bar\mu\) be the global mean. We compute

\[
  \Sigma_{\mathrm{task}}
  =
  \frac{1}{N}
  \sum_{c=1}^{C}
  n_c
  (\mu_c-\bar\mu)
  (\mu_c-\bar\mu)^\top.
\]
If \(U\) contains the top-\(r\) PCA directions computed from the same
prediction-position representations, the alignment coefficient is

\[
  \rho
  =
  \frac{
    \operatorname{Tr}
    \left(
      U^\top\Sigma_{\mathrm{task}}U
    \right)
  }{
    \operatorname{Tr}
    \left(
      \Sigma_{\mathrm{task}}
    \right)
  }.
\]
This diagnostic requires no gradients and is computed from one frozen forward
pass followed by covariance and trace-ratio calculations.

Table~\ref{tab:app-rho-gain} contains the seven values used for the main
correlation analysis. CoLA gains are measured in Matthews correlation points,
and all other gains are accuracy points. Across the seven tasks, the Pearson
correlation is \(r=0.93\). An exact two-sided permutation test evaluates all
\(7!=5{,}040\) assignments of the gain values to the fixed alignment values
and gives \(p=0.002\). 

\begin{table}[H]
\centering
\caption{Alignment \(\rho\) at \(N=64\) and PivotLoRA's shared-protocol gain
over vanilla LoRA.}
\label{tab:app-rho-gain}
\small
\begin{tabular}{@{}lccc@{}}
\toprule
Task & $\rho$ & Fisher ratio & PivotLoRA gain (points) \\
\midrule
AG~News & 0.843 & 0.14 & +19.4 \\
SST-2   & 0.667 & 0.02 & +14.3 \\
CoLA    & 0.530 & 0.02 & +12.5 \\
QNLI    & 0.458 & 0.02 & +3.9 \\
MNLI    & 0.419 & 0.03 & +0.1 \\
BoolQ   & 0.402 & 0.01 & +1.6 \\
RTE     & 0.293 & 0.01 & +1.8 \\
\bottomrule
\end{tabular}
\end{table}
\FloatBarrier

The class-based definition of \(\rho\) requires discrete labels. It is not
directly defined for open-ended generated responses and is not used to predict
the reasoning gains.

\subsection{Robustness of the Alignment Diagnostic}
\label{app:rho-robustness}

The seven-task Spearman correlation is 0.86. Leave-one-task-out Pearson
correlations range from 0.87 to 0.96. Removing AG~News still gives
\(r=0.87\), so the association is not determined only by the largest-gain
task.

As a same-sample-bias check, we compute a cross-split estimate on AG~News,
SST-2, and CoLA. The PCA subspace is estimated using one half of the \(N=64\)
set, while \(\Sigma_{\mathrm{task}}\) is estimated using the other half. We
then exchange the halves and average the two estimates. The absolute values
decrease because each estimate uses fewer examples, but the ordering
AG~News \(>\) SST-2 \(>\) CoLA is preserved.

\begin{table}[H]
\centering
\caption{Cross-split robustness check at \(N=64\). CoLA gains are in Matthews
correlation points, and the other gains are in accuracy points.}
\label{tab:rho-cross-split}
\small
\begin{tabular}{@{}lccc@{}}
\toprule
Task & Original $\rho$ & Cross-split $\rho$ & PivotLoRA gain \\
\midrule
AG~News & 0.843 & 0.390 & +19.4 \\
SST-2   & 0.667 & 0.285 & +14.3 \\
CoLA    & 0.530 & 0.259 & +12.5 \\
\bottomrule
\end{tabular}
\end{table}
\FloatBarrier

\paragraph{Subset variability.}
To estimate the sampling uncertainty of \(\rho\), we recomputed it on ten
independently drawn class-balanced \(N=64\) subsets per task
(Table~\ref{tab:rho-subsets}). AG~News is consistently high, and SST-2 is high
in eight of ten subsets, whereas QNLI, MNLI, RTE, and BoolQ exceed \(0.6\) in
only one subset each. In this BERT-base, rank-8, \(N=64\) configuration, values
near or above \(0.6\) therefore separate the two consistently high-alignment
tasks from the low-gain group.  CoLA, at \(\rho=0.53\), lies between the two groups on both alignment and gain.

\begin{table}[H]
\centering
\caption{\(\rho\) over ten class-balanced \(N=64\) subsets per task, with the
shared-protocol PivotLoRA gain from Table~\ref{tab:app-rho-gain}. CoLA was not included in this resampling.}
\label{tab:rho-subsets}
\small
\begin{tabular}{@{}lcccc@{}}
\toprule
Task & Mean \(\pm\) SD & Range & Subsets with \(\rho\ge0.6\) & PivotLoRA gain (points) \\
\midrule
AG~News & 0.843 \(\pm\) 0.017 & 0.823--0.874 & 10/10 & +19.4 \\
SST-2   & 0.673 \(\pm\) 0.185 & 0.191--0.824 & 8/10  & +14.3 \\
QNLI    & 0.440 \(\pm\) 0.079 & 0.360--0.623 & 1/10  & +3.9 \\
BoolQ   & 0.457 \(\pm\) 0.137 & 0.306--0.749 & 1/10  & +1.6 \\
MNLI    & 0.444 \(\pm\) 0.120 & 0.289--0.647 & 1/10  & +0.1 \\
RTE     & 0.391 \(\pm\) 0.109 & 0.255--0.606 & 1/10  & +1.8 \\
\bottomrule
\end{tabular}
\end{table}
\FloatBarrier

\paragraph{Class imbalance.}
On AG~News we drew 30 \(N=64\) subsets under each of three class-count profiles.
Balanced sampling (16/16/16/16) gives \(\rho=0.869\pm0.027\), moderate imbalance
(28/16/12/8) gives \(0.866\pm0.024\), and severe imbalance (40/12/8/4) gives
\(0.818\pm0.028\), with a minimum of \(0.755\). Severe imbalance lowers the
estimate but does not change the high-alignment conclusion in this experiment.
We did not test label noise. Because \(\rho\) is built from class means, noisy
labels can distort it directly.

For practical use, we recommend estimating uncertainty by recomputing \(\rho\)
over class-stratified resamples of the cached activations. A low or unstable
estimate should be treated as weak evidence for direction-aware
initialization, and method selection should still be validated under matched
tuning when feasible. We do not recommend a universal threshold. Instead, \(\rho\) should be treated as a
comparative screening diagnostic under a fixed model, rank, representation,
and sampling protocol.

\subsection{Encoder-Specific Class Structure}
\label{app:cls-structure}

Table~\ref{tab:app-cls-structure} reports BERT-specific measurements from
pretrained \texttt{[CLS]} representations. Fisher's ratio summarizes overall
class separation, while \(\rho\) measures the fraction of between-class
scatter captured by the top-\(r\) PCA subspace. These quantities need not
produce identical task orderings because one measures the amount of
between-class separation and the other measures its orientation relative to
the leading covariance directions.

\begin{table}[H]
\centering
\caption{Class structure in frozen BERT \texttt{[CLS]} representations,
averaged over three seeds. Columns denote the total number of labeled
examples.}
\label{tab:app-cls-structure}
\small
\begin{tabular}{@{}lccccc@{}}
\toprule
& $N=16$ & $N=32$ & $N=64$ & $N=128$ & $N=256$ \\
\midrule
\multicolumn{6}{@{}l}{\emph{Fisher's discriminant ratio}
\(\operatorname{tr}(S_B)/\operatorname{tr}(S_W)\)} \\
\midrule
SST-2   & 0.07 & 0.04 & 0.02 & 0.01 & 0.01 \\
CoLA    & 0.09 & 0.05 & 0.02 & 0.01 & 0.01 \\
AG~News & 0.37 & 0.18 & 0.14 & 0.11 & 0.10 \\
MNLI    & 0.16 & 0.07 & 0.03 & 0.02 & 0.01 \\
RTE     & 0.07 & 0.03 & 0.01 & 0.01 & 0.00 \\
QNLI    & 0.07 & 0.04 & 0.02 & 0.01 & 0.01 \\
\midrule
\multicolumn{6}{@{}l}{\emph{PCA-class alignment \(\rho\)}} \\
\midrule
SST-2   & 0.803 & 0.653 & 0.673 & 0.495 & 0.582 \\
CoLA    & 0.824 & 0.721 & 0.530 & 0.403 & 0.566 \\
AG~News & 0.911 & 0.821 & 0.843 & 0.876 & 0.901 \\
MNLI    & 0.760 & 0.536 & 0.423 & 0.310 & 0.351 \\
RTE     & 0.705 & 0.478 & 0.299 & 0.345 & 0.280 \\
QNLI    & 0.756 & 0.560 & 0.453 & 0.424 & 0.437 \\
\bottomrule
\end{tabular}
\end{table}
\FloatBarrier

Table~\ref{tab:app-cls-structure} comes from a separate run of the class-structure analysis that drew its few-shot subsets independently, so its \(N=64\) column differs from Table~\ref{tab:app-rho-gain} by at most 0.006 (0.673 versus 0.667 on SST-2). The correlation analysis uses Table~\ref{tab:app-rho-gain}.

\subsection{Per-Setting Gains Behind the Regime Map}
\label{app:regime-summary}

Table~\ref{tab:regime-detail} gives the range of per-setting gains behind
Table~\ref{tab:regime}. BERT rows cover the six primary tasks. Llama
classification rows cover SST-2 and AG~News on Llama-2-7B and Llama-3.1-8B and
BoolQ on Llama-3.1-8B. The reasoning row covers GSM8K and StrategyQA on
Llama-3.1-8B at \(N=128\) and evaluates the complete response-position
procedure (Appendix~\ref{app:reasoning-details}). The GPT-2 row covers XSum and CNN/DailyMail at \(N\in\{64,256\}\) with response positions and the single learning rate (Appendix~\ref{app:generation}).

\begin{table}[H]
\centering
\caption{PivotLoRA gain over vanilla LoRA by regime, in points of each task's
primary metric (ROUGE-L for GPT-2).}
\label{tab:regime-detail}
\small
\begin{tabular}{@{}lcccc@{}}
\toprule
Setting & \(d\) & \(d/N\) & Mean gain & Range \\
\midrule
BERT-base, \(N\le64\)        & 768  & 12--48     & \(+6.9\) & \(+0.1\) to \(+23.9\) \\
BERT-base, \(N\ge128\)       & 768  & 3--6       & \(+3.2\) & \(-2.0\) to \(+10.1\) \\
BERT-base, full data         & 768  & 0.002--0.31 & \(+0.4\) & \(-0.9\) to \(+2.2\) \\
Llama classification, \(N=64\)  & 4096 & 64 & \(+3.8\) & \(+0.5\) to \(+6.1\) \\
Llama classification, \(N=256\) & 4096 & 16 & \(+0.4\) & \(-1.8\) to \(+4.2\) \\
Llama-3.1-8B reasoning       & 4096 & 32         & \(+4.1\) & \(+3.6\) to \(+4.5\) \\
GPT-2 summarization          & 768  & 3--12      & \(+0.3\) & \(-0.1\) to \(+0.8\) \\
\bottomrule
\end{tabular}
\end{table}
\FloatBarrier

\subsection{Data-Efficiency Illustration}
\label{app:data-efficiency}

\begin{figure}[H]
\centering
\includegraphics[
  width=0.65\columnwidth
]{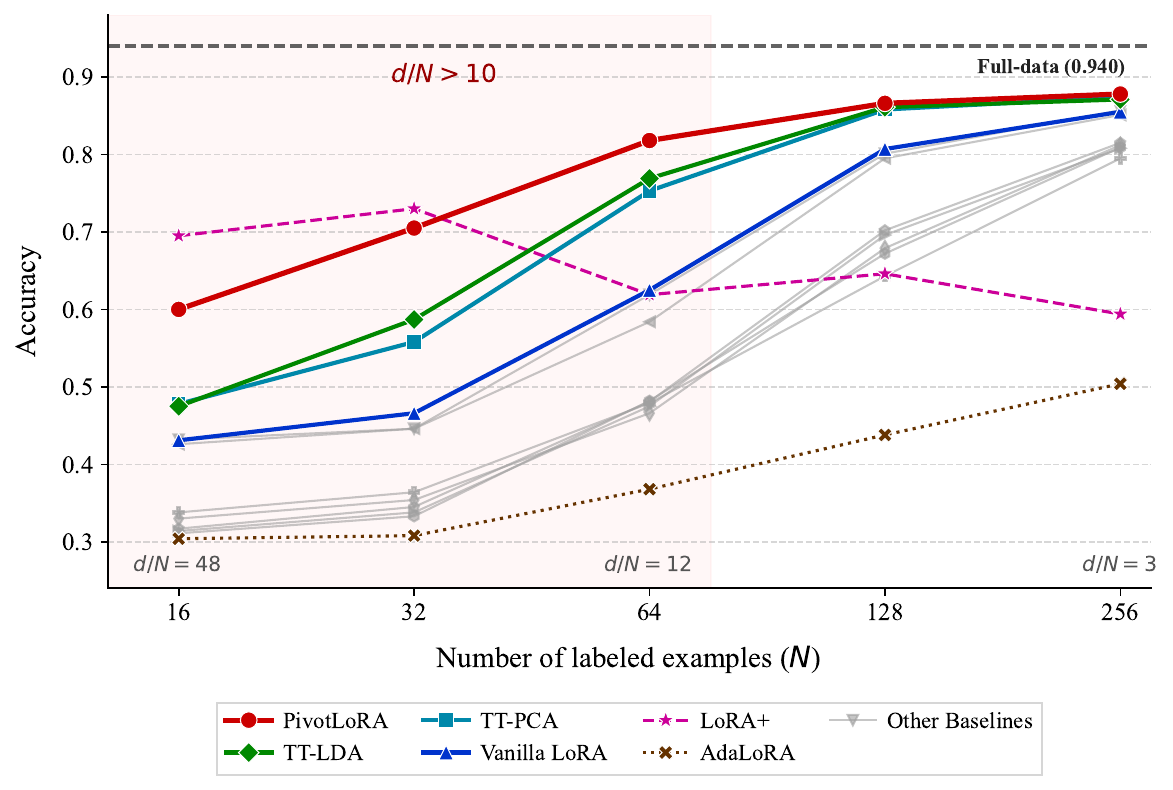}
\caption{Data efficiency on AG~News with BERT-base. The shaded region marks
the high-\(d/N\) few-shot regime. The dashed line denotes the displayed
full-data accuracy reference.}
\label{fig:app-efficiency}
\end{figure}

Figure~\ref{fig:app-efficiency} provides a task-specific illustration on
AG~News. At \(N=64\), PivotLoRA reaches approximately 89\% of the displayed
full-data accuracy ceiling, compared with approximately 68\% for vanilla
LoRA. 

\FloatBarrier

\subsection{Activation-Source Ablation}
\label{app:activation-source}

We compare activation sources for estimating the PivotLoRA initialization
subspace. Target-position initialization uses one \texttt{[CLS]} activation
per example, whereas all-token initialization uses all non-padding token
activations. The first two configurations use the same source for query and value adapters. The last two use the target position for query adapters and either all tokens or attention-weighted pooling for value adapters. All
other components, including the sampled examples, rank, scale, learning-rate
allocation, pivot timing, optimizer-state policy, and training budget, are
held fixed within each paired comparison.

\begin{table}[t]
\centering
\caption{Activation-source ablation for BERT-base PivotLoRA at $N=64$.
Values are mean $\pm$ sample standard deviation over three paired seeds.
Paired gains are computed as all-token minus target-position accuracy.
The last two rows use different sources for query and value adapters.}
\label{tab:activation-source}
\small
\begin{tabular}{@{}lcc@{}}
\toprule
Activation source & AG~News & SST-2 \\
\midrule
Target position (\texttt{[CLS]})
& .814$\pm$.015 & .778$\pm$.063 \\
All non-padding tokens
& .845$\pm$.008 & .802$\pm$.048 \\
Target for query, all tokens for value
& .835$\pm$.007 & .784$\pm$.024 \\
Target for query, attention-weighted for value
& .813$\pm$.017 & .666$\pm$.045 \\
\midrule
Paired gain of all-token
& +.032 & +.023 \\
All-token seed-level wins
& 3/3 & 2/3 \\
\bottomrule
\end{tabular}
\end{table}

This ablation is a separate matched sweep, so its absolute values should be compared within Table~\ref{tab:activation-source}.

Under this matched comparison, all-token initialization has a higher mean by
3.15 points on AG~News and 2.33 points on SST-2, with seed-level
wins in three of three and two of three comparisons, respectively.  The SST-2 difference is small relative to seed variation.

Assigning the target position to query adapters and all tokens to value
adapters also improves on the uniform target-only configuration on AG~News
(+2.1 points, 3/3 seeds) and slightly on SST-2 (+0.6 points, 1/3 seeds),
showing that sources can be selected per projection type. The tested
attention-weighted value source is neutral on AG~News ($-0.1$ points) and
lower on SST-2 ($-11.2$ points), so distributed-token pooling is not
automatically beneficial.

Target-position PCA uses exactly 64 activation vectors per layer, whereas
all-token PCA uses about 3,400 vectors on AG~News and 830 on SST-2. The corresponding top-eight explained-variance ratios, averaged over the inputs of the twelve adapted layers (Appendix~\ref{app:spiked-validation} reports the final layer), are 55.88\%
and 67.75\% for target-position PCA, compared with 15.14\% and 21.16\% for
all-token PCA. 
More broadly, the comparison illustrates the alignment--estimation trade-off. Prediction-position pooling targets the representation used for supervision,
whereas all-token pooling estimates its subspace from a substantially larger
activation sample.

\section{Robustness and Ablation Studies}
\label{app:ablation}

Several analyses below were run as separate sweeps under the primary protocol. Their absolute scores differ slightly from the primary tables, so each is compared within its own table.

\subsection{Aggregate Paired Robustness Analysis}
\label{app:significance}

Each individual task--$N$ setting contains only three seeds, so we do not
emphasize per-setting significance tests. Instead, we summarize the direction
of the mean differences across the 30 primary BERT settings.

PivotLoRA exceeds vanilla LoRA in 29 of the 30 settings, with the only negative
difference at RTE \(N=128\). A two-sided exact sign test on these 30
setting-level comparisons gives

\[
  p=5.8\times10^{-8}
\]
under a null model in which positive and negative differences are equally
likely. Because the settings share a backbone and tasks, they are not independent, and the test summarizes the benchmark without a strict inferential guarantee.
The five-seed experiments below provide complementary seed-level evidence for
selected settings.

\subsection{Five-Seed Confirmation}
\label{app:5seed}

All seven bootstrap intervals for the paired gain exclude zero, with mean gains from $+0.032$ on QNLI to $+0.231$ on AG~News at $N=32$. We reran seven representative BERT settings with seeds 42--46.
Table~\ref{tab:app-5seed} reports the mean, standard deviation, and bootstrap
95\% confidence interval for the mean paired difference. The intervals were
computed from seed-aligned LoRA and PivotLoRA runs.

\begin{table}[H]
\centering
\caption{Five-seed confirmation on representative BERT-base settings using
seeds 42--46. The confidence interval is a bootstrap interval for the mean
paired difference, PivotLoRA minus LoRA.}
\label{tab:app-5seed}
\small
\begin{tabular}{@{}llcccc@{}}
\toprule
Task & $N$ & LoRA & PivotLoRA & $\Delta$ & 95\% CI \\
\midrule
AG~News & 16
& .397$\pm$.044 & .577$\pm$.035 & +.180 & [+.145, +.217] \\
AG~News & 32
& .490$\pm$.016 & .721$\pm$.047 & +.231 & [+.191, +.265] \\
AG~News & 64
& .650$\pm$.032 & .816$\pm$.022 & +.166 & [+.128, +.205] \\
SST-2 & 32
& .525$\pm$.026 & .615$\pm$.064 & +.090 & [+.056, +.125] \\
SST-2 & 64
& .655$\pm$.032 & .790$\pm$.022 & +.135 & [+.112, +.156] \\
CoLA & 64
& .148$\pm$.061 & .265$\pm$.095 & +.117 & [+.070, +.169] \\
QNLI & 64
& .564$\pm$.035 & .596$\pm$.037 & +.032 & [+.010, +.047] \\
\bottomrule
\end{tabular}
\end{table}
\FloatBarrier

\subsection{Fisher-Gradient Baseline}
\label{app:lora-da}

We implement a Fisher-gradient variant inspired by LoRA-DA~\citep{zhang2026lorada}. For per-example gradients \(g_i\), it
estimates

\[
  F_{\mathrm{diag}}
  =
  \frac{1}{N}
  \sum_{i=1}^{N}
  g_i\odot g_i
\]
and performs an SVD of the Fisher-weighted mean gradient

\[
  \bar G\odot\sqrt{F_{\mathrm{diag}}}.
\]
Tables~\ref{tab:app-lorada-agnews} and
\ref{tab:app-lorada-sst2} report the resulting scores, with vanilla LoRA rerun in the same sweep. LoRA-DA is below vanilla LoRA in all 10 settings, by 0.006 to 0.131.

\begin{table}[H]
\centering
\caption{LoRA-DA on AG~News. Values are accuracy means and standard
deviations over three seeds. LoRA is rerun in the same sweep.}
\label{tab:app-lorada-agnews}
\small
\begin{tabular}{@{}lccccc@{}}
\toprule
Method & 16 & 32 & 64 & 128 & 256 \\
\midrule
LoRA
& .410$\pm$.029 & .477$\pm$.078 & .624$\pm$.054
& .774$\pm$.033 & .858$\pm$.010 \\
LoRA-DA
& .317$\pm$.034 & .352$\pm$.015 & .493$\pm$.056
& .679$\pm$.066 & .814$\pm$.030 \\
\bottomrule
\end{tabular}
\end{table}

\begin{table}[H]
\centering
\caption{LoRA-DA on SST-2. Values are accuracy means and standard deviations
over three seeds. LoRA is rerun in the same sweep.}
\label{tab:app-lorada-sst2}
\small
\begin{tabular}{@{}lccccc@{}}
\toprule
Method & 16 & 32 & 64 & 128 & 256 \\
\midrule
LoRA
& .541$\pm$.017 & .527$\pm$.026 & .639$\pm$.063
& .777$\pm$.010 & .847$\pm$.007 \\
LoRA-DA
& .528$\pm$.016 & .510$\pm$.020 & .598$\pm$.028
& .702$\pm$.056 & .841$\pm$.001 \\
\bottomrule
\end{tabular}
\end{table}
\FloatBarrier

\subsection{LoRA+ Behavior at Low Labeled-Example Counts}
\label{app:loraplus}

LoRA+ has the best mean in 5 of the 18 primary settings with \(N\le64\), all at \(N\le32\), where activation-based estimates are noisiest. It retains a random LoRA input basis and changes only the relative learning rates of the two factors.

Under the shared learning-rate protocol, LoRA+ collapses to a constant prediction on most tasks at \(N\ge128\) (zero standard deviation across seeds), and its AG~News scores vary widely.  Appendix~\ref{app:lr-sweep} includes a separate learning-rate and differential-ratio sensitivity analysis.

\subsection{Pivot Timing}
\label{app:ablation-pivot}

We vary the pivot epoch over

\[
  K\in\{1,2,3,5,7,10\}
\]
on SST-2, CoLA, and AG~News, with vanilla LoRA rerun in the same sweep.  Every tested pivot epoch outperforms LoRA at every \(N\le64\) on all three tasks, and the default \(K=3\) is competitive but not uniformly optimal. An early pivot allows more training in the orthonormal
coordinates, while a later pivot permits more task-gradient drift before the
coordinate change. 

\begin{table}[H]
\centering
\caption{Pivot timing on SST-2. Values are accuracy means over three seeds.}
\label{tab:app-pivot-sst2}
\small
\begin{tabular}{@{}lccccc@{}}
\toprule
$K$ & 16 & 32 & 64 & 128 & 256 \\
\midrule
1
& .570 & .601 & .762 & .818 & .854 \\
2
& .568 & .630 & .779 & .821 & .854 \\
3 (default)
& .567 & .608 & .786 & .829 & .844 \\
5
& .565 & .619 & .794 & .825 & .851 \\
7
& .569 & .659 & .797 & .820 & .851 \\
10
& .561 & .612 & .783 & .813 & .850 \\
\midrule
LoRA
& .543 & .526 & .641 & .777 & .846 \\
\bottomrule
\end{tabular}
\end{table}

\begin{table}[H]
\centering
\caption{Pivot timing on CoLA. Values are Matthews correlation means over
three seeds.}
\label{tab:app-pivot-cola}
\small
\begin{tabular}{@{}lccccc@{}}
\toprule
$K$ & 16 & 32 & 64 & 128 & 256 \\
\midrule
1
& .127 & .183 & .285 & .403 & .412 \\
2
& .126 & .191 & .296 & .378 & .400 \\
3 (default)
& .119 & .182 & .289 & .366 & .393 \\
5
& .132 & .172 & .276 & .355 & .388 \\
7
& .128 & .183 & .279 & .365 & .403 \\
10
& .125 & .175 & .270 & .369 & .401 \\
\midrule
LoRA
& .100 & .113 & .200 & .339 & .400 \\
\bottomrule
\end{tabular}
\end{table}

\begin{table}[H]
\centering
\caption{Pivot timing on AG~News. Values are accuracy means over three seeds.}
\label{tab:app-pivot-agnews}
\small
\begin{tabular}{@{}lccccc@{}}
\toprule
$K$ & 16 & 32 & 64 & 128 & 256 \\
\midrule
1
& .589 & .708 & .799 & .857 & .877 \\
2
& .593 & .717 & .802 & .861 & .878 \\
3 (default)
& .604 & .707 & .816 & .866 & .878 \\
5
& .595 & .724 & .817 & .865 & .879 \\
7
& .583 & .719 & .821 & .865 & .877 \\
10
& .516 & .701 & .813 & .865 & .879 \\
\midrule
LoRA
& .411 & .477 & .631 & .774 & .857 \\
\bottomrule
\end{tabular}
\end{table}
\FloatBarrier

\subsection{Pivot Preservation and Orthogonality Diagnostics}
\label{app:pivot-diagnostics}

For each classification adapter, we measure the relative preservation error

\[
  \epsilon_{\mathrm{pivot}}
  =
  \frac{
    \left\|
      B^{(K)}A^{(K)}
      -
      B^{\mathrm{new}}A^{\mathrm{new}}
    \right\|_F
  }{
    \left\|
      B^{(K)}A^{(K)}
    \right\|_F
    +
    10^{-12}
  }.
\]
The factor \(\alpha/r\) cancels from this ratio. We also measure the residual
between the old and new row spans, the row-orthogonality error

\[
  \epsilon_{\mathrm{orth}}(A)
  =
  \|AA^\top-I_r\|_F,
\]
and the change in the layerwise alignment score. The classification values in
Table~\ref{tab:pivot-diagnostics} are aggregated over three seeds and 72
adapter instances per task, comprising 24 adapted modules for each seed.

\begin{table}[H]
\centering
\caption{Classification output-preservation and orthogonality diagnostics.
The alignment change excludes layer 0 because its embedding-level
\texttt{[CLS]} scatter is near zero.}
\label{tab:pivot-diagnostics}
\resizebox{\textwidth}{!}{%
\renewcommand{\arraystretch}{1.25}
\begin{tabular}{@{}lcccccc@{}}
\toprule
Task
& Mean $\epsilon_{\mathrm{pivot}}$
& Max.\ $\epsilon_{\mathrm{pivot}}$
& Mean span residual
& $\epsilon_{\mathrm{orth}}(A^{(K)})$
& $\epsilon_{\mathrm{orth}}(A^{\mathrm{new}})$
& Mean $\Delta\rho$ \\
\midrule
AG~News
& $3.14\times10^{-7}$
& $8.82\times10^{-7}$
& $2.94\times10^{-7}$
& 2.809
& $4.84\times10^{-7}$
& $2.77\times10^{-9}$ \\
CoLA
& $3.00\times10^{-7}$
& $5.07\times10^{-7}$
& $2.87\times10^{-7}$
& 2.809
& $4.69\times10^{-7}$
& $-2.31\times10^{-9}$ \\
SST-2
& $2.79\times10^{-7}$
& $7.91\times10^{-7}$
& $2.89\times10^{-7}$
& 2.809
& $4.76\times10^{-7}$
& $2.56\times10^{-10}$ \\
\bottomrule
\end{tabular}%
}
\end{table}
\FloatBarrier

The output and span residuals remain below \(10^{-6}\). The post-pivot
orthogonality errors are also below \(10^{-6}\), and the alignment change is
near numerical zero. These measurements confirm that the classification pivot preserves the represented update and learned row span while changing their coordinate parameterization. In the reasoning runs, the maximum relative preservation error is $1.44\times10^{-7}$ and the maximum post-pivot orthogonality error is $3.51\times10^{-8}$.

\paragraph{What the pivot changes.}
Writing $\sigma_1,\dots,\sigma_r$ for the singular values of $A^{(K)}$,
$\epsilon_{\mathrm{orth}}(A^{(K)})^2=r-2\sum_i\sigma_i^2+\sum_i\sigma_i^4$.
When $\|A^{(K)}\|_F^2=\sum_i\sigma_i^2$ is also logged, as in the reasoning
runs, these two quantities determine the effective rank
$r_{\mathrm{eff}}=(\sum_i\sigma_i^2)^2/\sum_i\sigma_i^4$ of the pre-pivot Gram
matrix. In classification, the value $2.809$ is what a scaled orthonormal
matrix $\sqrt{5/d}\,U^\top$ would give ($2.810$ for $d=768$), which places
$\|A^{(K)}\|_F$ within about 5\% of its initial value. The pivot therefore raises the row norms of $A$ about twelvefold, although the three-decimal value cannot resolve whether $A$ also became anisotropic. In the reasoning runs, the median
adapter's $\|A^{(K)}\|_F$ grows to $5.7\times$ its initial value, and the
median $r_{\mathrm{eff}}$ is $1.9$--$2.1$ on GSM8K and $1.4$--$1.5$ on
StrategyQA across seeds, out of $r=8$. Early training therefore concentrates
the learned input basis on a few directions, and the pivot restores an
isotropic basis for the same span.

\subsection{Orthogonal Pivot and Optimizer-State Ablation}
\label{app:ablation-reset}

We separate the output-preserving QR pivot from the classification
optimizer-state policy. \textsc{QR-Pivot-NoReset} applies the pivot but keeps the Adam state of \(A\), whereas \textsc{PivotLoRA} resets it. Both use the same post-pivot learning-rate handling (Appendix~\ref{app:classification-lr}).

\begin{table}[H]
\centering
\caption{Classification optimizer-state ablation on SST-2. Values are
accuracy means over three seeds.}
\label{tab:ablation_reset_sst2}
\small
\begin{tabular}{@{}lccccc@{}}
\toprule
Method & $N=16$ & $N=32$ & $N=64$ & $N=128$ & $N=256$ \\
\midrule
TT-PCA
& .520 & .530 & .755 & .817 & .858 \\
QR-Pivot-NoReset
& .552 & .623 & .795 & .829 & .858 \\
PivotLoRA
& .541 & .628 & .784 & .828 & .847 \\
\bottomrule
\end{tabular}
\end{table}

\begin{table}[H]
\centering
\caption{Classification optimizer-state ablation on CoLA. Values are
Matthews correlation means over three seeds.}
\label{tab:ablation_reset_cola}
\small
\begin{tabular}{@{}lccccc@{}}
\toprule
Method & $N=16$ & $N=32$ & $N=64$ & $N=128$ & $N=256$ \\
\midrule
TT-PCA
& .104 & .137 & .270 & .383 & .402 \\
QR-Pivot-NoReset
& .117 & .172 & .301 & .381 & .398 \\
PivotLoRA
& .117 & .174 & .304 & .381 & .393 \\
\bottomrule
\end{tabular}
\end{table}

\begin{table}[H]
\centering
\caption{Classification optimizer-state ablation on AG~News. Values are
accuracy means over three seeds.}
\label{tab:ablation_reset_agnews}
\small
\begin{tabular}{@{}lccccc@{}}
\toprule
Method & $N=16$ & $N=32$ & $N=64$ & $N=128$ & $N=256$ \\
\midrule
TT-PCA
& .478 & .561 & .756 & .857 & .874 \\
QR-Pivot-NoReset
& .589 & .689 & .811 & .859 & .878 \\
PivotLoRA
& .597 & .703 & .811 & .857 & .878 \\
\bottomrule
\end{tabular}
\end{table}
\FloatBarrier

The QR pivot accounts for most of the improvement over TT-PCA. Without the reset, it improves on TT-PCA in 12 of 15 settings, by $0.037$ on average and by up to $0.128$ on AG~News at $N=32$. The additional reset produces smaller, task-dependent differences and is not uniformly superior to
\textsc{QR-Pivot-NoReset}. We therefore treat the classification reset as a stabilization policy and not as the main source of the gain.

This ablation does not apply directly to the reasoning experiments, which
reset the optimizer states of both \(A\) and \(B\), and no reasoning
reset-only or no-reset ablation was conducted.

\subsection{Post-Pivot Learning-Rate Policy}
\label{app:post-pivot-lr}

The pivot's gain does not depend on how the \(B\) learning rate is handled afterward. All pivot variants share the TT-PCA initialization, the
eigenvalue-scaled rates before the pivot, the QR pivot at \(K=3\), and the Adam
reset for \(A\), and differ only in the \(B\) rate after the pivot. The
implemented policy raises a common rate from \(\eta_A\) to \(16\eta_A\) over
10 steps and then resumes the eigenvalue-scaled rates. The alternatives keep
the eigenvalue-scaled rates unchanged, switch to a uniform \(16\eta_A\) for the
rest of training, or ramp to that uniform rate and keep it. The sweep covers SST-2, CoLA, and AG~News at \(N\in\{16,32,64,128,256\}\) with three seeds.

\begin{table}[H]
\centering
\caption{Post-pivot learning-rate policy. Task columns average over the five
values of \(N\) (CoLA in Matthews correlation); ``Wins'' counts the 15
task--\(N\) settings in which a variant exceeds TT-PCA.}
\label{tab:post-pivot-lr}
\small
\begin{tabular}{@{}lccccc@{}}
\toprule
Variant & SST-2 & CoLA & AG~News & Average & Wins \\
\midrule
TT-PCA (no pivot) & .697 & .257 & .717 & .557 & -- \\
\midrule
Pivot, no rate change & .734 & .275 & .770 & .593 & 11/15 \\
Pivot, implemented policy & .728 & .276 & .772 & .592 & 12/15 \\
Pivot, uniform rate & .744 & .266 & .777 & .596 & 13/15 \\
Pivot, ramp then uniform rate & .723 & .265 & .766 & .585 & 12/15 \\
\bottomrule
\end{tabular}
\end{table}
\FloatBarrier

Every policy improves on TT-PCA by \(0.028\)--\(0.039\) on average (\(0.036\) with no rate change), and the policies differ by at most \(0.011\).

\subsection{Learning-Rate Structure in the Ablation Ladder}
\label{app:lr-structure}

TT-PCA and TT-LDA differ in both initialization direction and learning-rate
structure. TT-PCA uses the eigenvalue-scaled rates in
Eq.~\eqref{eq:eigenlr}, whereas TT-LDA uses the uniform differential rate
\(\eta_B=16\eta_A\). Their direct comparison is consequently an ablation ladder and does not isolate PCA from LDA.

Across the full BERT benchmark, TT-LDA, TT-PCA, and PivotLoRA have average
ranks 3.13, 3.77, and 1.83, respectively. LoRA+ retains random input
directions and applies a \(16\times\) differential rate but has average rank
8.43 under the shared protocol. This aggregate comparison argues against an
explanation based only on the average \(B\)-to-\(A\) learning-rate ratio. The direct controls in Appendix~\ref{app:direction-lr-controls} isolate the PCA directions from the eigenvalue-to-coordinate rate assignment.

\phantomsection
\label{app:lr-sweep}
\paragraph{Learning-rate sensitivity.}
We sweep learning rates on AG~News, SST-2, and CoLA at \(N=64\).

For each method, we sweep the base learning-rate anchor over

\[
  \eta_A\in
  \left\{
    5\times10^{-5},
    2\times10^{-4},
    8\times10^{-4}
  \right\}.
\]
The primary shared-protocol anchor is \(\eta_A=2\times10^{-4}\).
Differential-rate methods derive their \(B\)-factor rates from this anchor.
For LoRA+, model selection additionally ranges over

\[
  \eta_B/\eta_A\in\{4,8,16,32\}.
\]

Table~\ref{tab:lr-sensitivity} reports the average score at the primary anchor
and the average of each method's best swept score. Tuning reduces several margins, but PivotLoRA remains strongest on average
among the swept methods. 

\begin{table}[H]
\centering
\caption{Learning-rate sensitivity at \(N=64\). Scores are unweighted averages of each task's primary metric over AG~News, SST-2, and CoLA. ``Avg.\ anchor'' uses the primary base learning-rate anchor
\(\eta_A=2\times10^{-4}\), together with each method's prescribed factor-wise
allocation. ``Avg.\ best'' averages each method's best score over the base-rate
sweep. For LoRA+, selection additionally includes
\(\eta_B/\eta_A\in\{4,8,16,32\}\). Gaps are measured relative to PivotLoRA
under the corresponding comparison.}
\label{tab:lr-sensitivity}
\small
\begin{tabular}{@{}lcccc@{}}
\toprule
Method
& Avg.\ anchor
& Avg.\ best
& Pivot anchor gap
& Pivot best gap \\
\midrule
LoRA
& .522 & .622 & +.139 & +.045 \\
LoRA+
& .326 & .636 & +.335 & +.031 \\
PiSSA
& .420 & .606 & +.241 & +.061 \\
LoRAM
& .428 & .617 & +.233 & +.050 \\
EVA
& .424 & .618 & +.237 & +.048 \\
TT-PCA
& .643 & .655 & +.017 & +.012 \\
PivotLoRA
& .661 & .667 & .000 & .000 \\
\bottomrule
\end{tabular}
\end{table}

\begin{table}[H]
\centering
\caption{Best swept scores by task at \(N=64\). Each score is selected over
the base learning-rate anchors
\(\{5\times10^{-5},2\times10^{-4},8\times10^{-4}\}\). For LoRA+, selection
additionally ranges over the differential ratios
\(\eta_B/\eta_A\in\{4,8,16,32\}\).}
\label{tab:lr-best}
\small
\begin{tabular}{@{}lccc@{}}
\toprule
Method & AG~News & SST-2 & CoLA \\
\midrule
LoRA
& .781 & .784 & .301 \\
LoRA+
& .833 & .757 & .318 \\
PiSSA
& .754 & .733 & .330 \\
LoRAM
& .732 & .789 & .330 \\
EVA
& .742 & .786 & .327 \\
TT-PCA
& .817 & .821 & .328 \\
PivotLoRA
& .843 & .822 & .335 \\
\bottomrule
\end{tabular}
\end{table}
\FloatBarrier

\subsection{Direction versus Learning-Rate Scaling Controls}
\label{app:direction-lr-controls}

We evaluate three controls at \(N=64\) to separate target-token PCA directions
from the associated learning-rate allocation.

\begin{itemize}
  \item \textsc{Random+EigenLR} uses a random input basis drawn as in vanilla LoRA
  (Kaiming-uniform, not orthonormalized) with the same eigenvalue-scaled
  per-rank \(B\) learning rates.
  \item \textsc{TT-PCA+UniformLR} retains the target-token PCA basis and uses a
  uniform differential learning rate.
  \item \textsc{TT-PCA+ShuffledLR} retains the target-token PCA basis and
  randomly permutes the eigenvalue-to-coordinate rate assignment.
\end{itemize}

The TT-PCA row in Table~\ref{tab:direction-lr-controls} comes from the same control sweep and is the reference for the three variants.

\begin{table}[H]
\centering
\caption{Direction versus learning-rate scaling controls at \(N=64\) using
BERT-base. Values are mean $\pm$ standard deviation over three seeds from the
separate diagnostic sweep.}
\label{tab:direction-lr-controls}
\small
\setlength{\tabcolsep}{4pt}
\begin{tabular}{@{}lccc@{}}
\toprule
Method & AG~News & SST-2 & CoLA \\
\midrule
LoRA
& .640$\pm$.033 & .647$\pm$.040 & .188$\pm$.042 \\
TT-PCA
& .751$\pm$.025 & .769$\pm$.038 & .268$\pm$.092 \\
Random+EigenLR
& .444$\pm$.268 & .571$\pm$.031 & .028$\pm$.035 \\
TT-PCA+UniformLR
& .752$\pm$.019 & .769$\pm$.035 & .276$\pm$.100 \\
TT-PCA+ShuffledLR
& .754$\pm$.028 & .765$\pm$.028 & .281$\pm$.088 \\
\bottomrule
\end{tabular}
\end{table}
\FloatBarrier

A random basis does not become competitive merely by receiving the
eigenvalue-scaled rates. Conversely, the target-token PCA basis retains its
performance when the rates are made uniform or their assignments are
permuted. Within this control sweep, the selected directions therefore explain
more of the advantage than the eigenvalue-to-coordinate rate assignment.
Because the classification implementation applies a fixed random rotation
within the PCA span (Eq.~\eqref{eq:pca-init}), each rate-scaled row of $B$
already corresponds to a mixture of principal directions, so the shuffled control
therefore tests the rate allocation itself, not a pairing between individual
directions and rates. This experiment separates direction from learning-rate allocation but not from initialization norm.

\section{Per-Seed and Cost Details}
\label{app:efficiency}

\subsection{Per-Seed Decoder-Scale Results}
\label{app:perseed-tables}

Tables~\ref{tab:app-llama2-perseed} and
\ref{tab:app-llama31-perseed} provide the per-seed decoder-scale results.
PivotLoRA has the highest \(N=64\) aggregate mean on SST-2 and AG~News for
both backbones, although individual seed-level exceptions occur. 

\begin{table}[H]
\centering
\caption{Llama-2-7B per-seed classification results in bf16 on an NVIDIA L4.}
\label{tab:app-llama2-perseed}
\small
\setlength{\tabcolsep}{3pt}
\begin{tabular}{@{}l|ccc|ccc|ccc|ccc@{}}
\toprule
& \multicolumn{3}{c|}{SST-2, $N=64$}
& \multicolumn{3}{c|}{SST-2, $N=256$}
& \multicolumn{3}{c|}{AG~News, $N=64$}
& \multicolumn{3}{c}{AG~News, $N=256$} \\
& s42 & s43 & s44
& s42 & s43 & s44
& s42 & s43 & s44
& s42 & s43 & s44 \\
\midrule
LoRA
& .711 & .757 & .655
& .931 & .917 & .922
& .724 & .743 & .777
& .877 & .884 & .876 \\
PiSSA
& .683 & .571 & .650
& .911 & .927 & .930
& .706 & .709 & .684
& .867 & .874 & .854 \\
LoRAM
& .765 & .787 & .737
& .927 & .944 & .915
& .719 & .771 & .737
& .878 & .877 & .872 \\
PivotLoRA
& .802 & .742 & .764
& .927 & .921 & .889
& .809 & .791 & .753
& .876 & .869 & .839 \\
\bottomrule
\end{tabular}
\end{table}
\FloatBarrier

\begin{table}[H]
\centering
\caption{Llama-3.1-8B per-seed classification results in bf16 on an
NVIDIA L4.}
\label{tab:app-llama31-perseed}
\small
\setlength{\tabcolsep}{3pt}
\begin{tabular}{@{}l|ccc|ccc|ccc|ccc@{}}
\toprule
& \multicolumn{3}{c|}{SST-2, $N=64$}
& \multicolumn{3}{c|}{SST-2, $N=256$}
& \multicolumn{3}{c|}{AG~News, $N=64$}
& \multicolumn{3}{c}{AG~News, $N=256$} \\
& s42 & s43 & s44
& s42 & s43 & s44
& s42 & s43 & s44
& s42 & s43 & s44 \\
\midrule
LoRA
& .728 & .723 & .661
& .912 & .885 & .851
& .758 & .783 & .737
& .870 & .885 & .886 \\
PiSSA
& .626 & .578 & .663
& .911 & .865 & .853
& .710 & .744 & .691
& .865 & .874 & .854 \\
LoRAM
& .720 & .640 & .627
& .907 & .909 & .888
& .735 & .800 & .708
& .877 & .879 & .861 \\
PivotLoRA
& .735 & .778 & .614
& .888 & .912 & .899
& .800 & .799 & .798
& .889 & .884 & .836 \\
\bottomrule
\end{tabular}
\end{table}
\FloatBarrier

\subsection{Initialization Overhead}
\label{app:overhead}

Table~\ref{tab:overhead} reports initialization time and additional persistent
storage for representative methods. PivotLoRA performs one activation pass and
PCA initialization before training and one QR-based coordinate change at the
pivot. The timing is small relative to the approximately 69-second
classification training run used in the cost comparison below.

\begin{table}[H]
\centering
\caption{Initialization overhead for BERT-base with 12 layers and rank
\(r=8\) on an NVIDIA T4.}
\label{tab:overhead}
\small
\begin{tabular}{@{}lcc@{}}
\toprule
Method & Initialization time & Additional persistent storage \\
\midrule
LoRA
& $<0.1$\,s & None \\
PivotLoRA
& $<1$\,s plus $<1$\,s at pivot & None \\
PiSSA
& ${\sim}2$\,s, layerwise SVD & Residual \(W^{\mathrm{res}}\) \\
LoRA-GA
& ${\sim}3$\,s, gradient plus SVD & None \\
LoRA-One
& ${\sim}3$\,s, gradient plus SVD & None \\
EVA
& ${\sim}1$\,s, activation SVD & None \\
LoRAM
& $<0.5$\,s, power iteration & None \\
\bottomrule
\end{tabular}
\end{table}
\FloatBarrier

\subsection{Computational Cost}
\label{app:cost}

Tables~\ref{tab:app-cost-bert} and~\ref{tab:app-cost-llama} report
end-to-end training time and peak memory. On the BERT measurement,
PivotLoRA adds 4.0 seconds relative to LoRA, corresponding to 5.8\% wall-time
overhead. Most of this difference comes from the per-rank optimizer groups for the eigenvalue-scaled learning rates, and the QR pivot itself takes under a second. Peak memory is nearly unchanged.

\begin{table}[H]
\centering
\caption{BERT-base training time and peak memory at \(N=64\) on SST-2,
averaged over two seeds on one NVIDIA T4.}
\label{tab:app-cost-bert}
\small
\begin{tabular}{@{}lccc@{}}
\toprule
Method & Wall time (s) & Peak VRAM (MB) & Overhead vs.\ LoRA \\
\midrule
LoRA
& 69.4 & 1{,}266 & N/A \\
DoRA
& 69.5 & 1{,}399 & +0.1\,s (+0.2\%) \\
AdaLoRA
& 102.2 & 1{,}264 & +32.9\,s (+47.4\%) \\
PiSSA
& 68.6 & 1{,}267 & $-0.8$\,s ($-1.2\%$) \\
LoRA-GA
& 68.0 & 1{,}272 & $-1.4$\,s ($-1.9\%$) \\
LoRA-One
& 69.0 & 1{,}272 & $-0.4$\,s ($-0.6\%$) \\
LoRA+
& 68.9 & 1{,}267 & $-0.5$\,s ($-0.7\%$) \\
EVA
& 69.8 & 1{,}267 & +0.4\,s (+0.6\%) \\
LoRAM
& 68.9 & 1{,}267 & $-0.5$\,s ($-0.7\%$) \\
SRLoRA
& 89.4 & 1{,}267 & +20.0\,s (+28.8\%) \\
\midrule
TT-PCA
& 73.6 & 1{,}267 & +4.2\,s (+6.0\%) \\
TT-LDA
& 69.3 & 1{,}267 & $-0.1$\,s ($-0.1\%$) \\
PivotLoRA
& 73.4 & 1{,}268 & +4.0\,s (+5.8\%) \\
\bottomrule
\end{tabular}

\vspace{2pt}
\begin{minipage}{0.92\linewidth}
\footnotesize
TT-PCA and PivotLoRA use per-rank optimizer groups for their
eigenvalue-scaled rates. TT-LDA uses the uniform differential rate
\(\eta_B=16\eta_A\). The 0.2-second difference between TT-PCA and PivotLoRA
is within run-to-run timing variation.
\end{minipage}
\end{table}
\FloatBarrier

\begin{table}[H]
\centering
\caption{Llama-2-7B classification training cost on SST-2 at \(N=64\), using
bf16 on one NVIDIA L4.}
\label{tab:app-cost-llama}
\small
\begin{tabular}{@{}lcc@{}}
\toprule
Method & Wall time (s) & Peak VRAM (GB) \\
\midrule
LoRA
& ${\sim}74$ & 14.6 \\
PiSSA
& ${\sim}73$ & 14.6 \\
PivotLoRA
& ${\sim}73$ & 14.6 \\
\bottomrule
\end{tabular}
\end{table}
\FloatBarrier

\section{Limitations}
\label{app:limitations}

\paragraph{Activation-source and task scope.}
Activation-based initialization works best with a source matched to the
supervision, such as the \texttt{[CLS]} representation for BERT
classification, the sequence mean for decoder classification, and hidden
states that predict supervised response tokens for completion-only reasoning.
Target-position pooling emphasizes representations tied to prediction, whereas
all-token pooling provides more observations but includes variation less
specific to the target, and neither should be expected to dominate across
tasks. Query and value adapters share one source rule in our main
configuration. Sources can be chosen per projection type
(Appendix~\ref{app:activation-source}), but we do not study this
systematically.

\paragraph{Theoretical and diagnostic scope.}
The coefficient \(\rho\) is defined through between-class covariance, so it
applies most directly to labeled classification, and its association with
gains is measured across seven BERT tasks under one protocol. Generation and
continuous targets would require a different definition of supervised
structure. The perturbation result assumes independent sub-Gaussian
observations, an approximately spiked covariance, and a positive eigengap, and
the first-step analysis adds a class-conditional mean model and an assumption
relating gradient-weighted geometry to between-class scatter. Both give a
regime-level account and are not convergence guarantees for transformer
fine-tuning.

\paragraph{Model and task coverage.}
The decoder classification experiments cover fewer tasks and baselines than
the BERT study and check whether the main pattern persists at larger scale,
not full rankings of methods for 7B and 8B models. The reasoning experiments
cover mathematical and yes/no reasoning with one 8B model and do not support
claims about instruction tuning, multilingual adaptation, or unrestricted
generation.

\paragraph{Reasoning attribution and statistical scope.}
The reasoning procedure combines response-position PCA, a fixed
\(\sqrt{5/d}\) basis scale, one output-preserving pivot, and a reset of the
AdamW state for both factors. Without norm-matched, reset-only, or
direction-versus-rate controls in this setting, the results evaluate the
complete procedure and do not isolate subspace direction as the cause. The
McNemar tests in Appendix~\ref{app:reasoning-results} treat test items as the
unit of variation and do not capture variation across training runs.

\paragraph{Pivot scope.}
The pivot orthonormalizes coordinates within the row span learned before
epoch \(K\) and adds no new directions. Replacing low-utility rank directions
during training while preserving the accumulated update is a natural
extension.

\paragraph{Broader impacts.}
Parameter-efficient adaptation lowers the cost of few-shot specialization,
which may also lower the cost of adapting pretrained models for harmful or
deceptive uses. PivotLoRA should be used under the same safety,
data-governance, and misuse-mitigation practices that apply to the underlying
pretrained models.

\FloatBarrier

\end{document}